\documentclass[a4paper,10pt]{article}

\usepackage{fullpage}
\usepackage[utf8]{inputenc} %
\usepackage[T1]{fontenc}    %
\usepackage[hidelinks]{hyperref}       %
\usepackage{url}            %
\usepackage{booktabs}       %
\usepackage{amsfonts}       %
\usepackage{nicefrac}       %
\usepackage{microtype}      %
\usepackage{xcolor}         %
\usepackage{Definitions}
\usepackage{algcompatible}
\usepackage[pdftex]{graphicx}
\usepackage{enumerate}
\usepackage{subfig}
\usepackage{comment}
\usepackage[export]{adjustbox}
\usepackage{mathtools}
\usepackage{caption}
\usepackage{listings}

\definecolor{codegreen}{rgb}{0,0.6,0}
\definecolor{codegray}{rgb}{0.5,0.5,0.5}
\definecolor{codepurple}{rgb}{0.58,0,0.82}
\definecolor{backcolour}{rgb}{0.95,0.95,0.92}

\lstdefinestyle{mystyle}{
    commentstyle=\color{codegreen},
    keywordstyle=\color{magenta},
    numberstyle=\tiny\color{codegray},
    stringstyle=\color{codepurple},
    basicstyle=\ttfamily\footnotesize,
    breakatwhitespace=false,
    breaklines=true,
    captionpos=b,
    keepspaces=true,
    numbersep=5pt,
    showspaces=false,
    showstringspaces=false,
    showtabs=false,
    tabsize=2,
    frame=single
}

\DeclareCaptionFormat{cont}{#1 (cont.)#2#3\par}

\usepackage[sort&compress,numbers]{natbib}

\title{Prediction-Powered Ranking of Large Language Models}
\author{Ivi Chatzi \and Eleni Straitouri \and Suhas Thejaswi \and Manuel Gomez-Rodriguez}

\date{
	Max Planck Institute for Software Systems \\ Kaiserslautern, Germany \\ \texttt{\{ichatzi,estraitouri,thejaswi,manuelgr\}@mpi-sws.org}\\
}

\begin{document}
\maketitle

\begin{abstract}
Large language models are often ranked according to their level of alignment with human pre\-fe\-ren\-ces---a model is better than other models if its outputs are more frequently preferred by humans. One of the popular ways to elicit human preferences utilizes pairwise comparisons between the outputs provided by diffe\-rent models to the same inputs. However, since gathering pairwise comparisons by humans is costly and time-consuming, it has become a common practice to gather pairwise comparisons by a strong large language model---a model strongly aligned with human preferences. Sur\-prising\-ly, practitioners cannot currently measure the uncertainty that any mismatch between human and model preferences may introduce in the constructed rankings. In this work, we develop a statistical framework to bridge this gap. Given a (small) set of pairwise comparisons by humans and a large set of pairwise comparisons by a model, our framework provides a rank-set---a set of possible ranking positions---for each of the models under comparison. Moreover, it guarantees that, with a probability greater than or equal to a user-specified value, the rank-sets cover the true ranking consistent with the distribution of human pairwise preferences asymptotically. Using pairwise comparisons made by humans in the LMSYS Chatbot Arena platform and pairwise comparisons made by three strong large language models, we empirically demonstrate the effectivity of our framework and show that the rank-sets constructed using only pairwise comparisons by the strong large language models are often inconsistent with (the distribution of) human pairwise preferences.

\vspace{2mm}

\end{abstract}

\section{Introduction}
\label{sec:introduction}
During the last years, large language models (LLMs) have shown a remarkable ability to 
generate and understand general-purpose language~\cite{bubeck2023sparks}.
As a result, there has been an increasing excitement in their potential to help
humans solve a variety of open-ended, complex tasks across many application domains
such as coding~\cite{mozannar2022reading}, healthcare~\cite{haupt2023ai} and scientific discovery~\cite{romera2023mathematical}, to name a few.
However, evaluating and comparing the performance of different LLMs has become 
very challenging~\cite{chang2024asurvey}.
The main reason is that, 
in contrast to traditional machine learning models, 
LLMs can solve a large number of different tasks and, in many of these tasks, there 
is not a unique, structured solution. 
As a consequence, there has been a paradigm shift towards evaluating their performance 
according to their level of alignment with human preferences---a model is better 
than other models if its outputs are more frequently preferred by humans~\cite{hendryckstest2021,wang2022self,ouyang2022training,wang2023aligning,lmsys2023chatbot}.

One of the most popular paradigms to rank a set of LLMs according to their level of alignment with human preferences utilizes pairwise comparisons~\cite{lmsys2023chatbot,taori2023stanford,zheng2023judging,li2023generative,li2023prd,boubdir2023elo,singhal2023large,chiang2024chatbot}. 
Under this paradigm, each pairwise comparison comprises the outputs of two different models picked uniformly at random to an input
sampled from a given dis\-tri\-bu\-tion of inputs.
Moreover, the pairwise comparisons are used to rank the models with a variety of methods such as the Elo
rating~\cite{askell2021general,dettmers2024qlora,bai2022training,wu2023chatarena,lin2023llm}, the Bradley-Terry model~\cite{boyeau2024autoeval,chiang2024chatbot,lmsys2023chatbot} or the win-rate~\cite{zheng2023judging,boyeau2024autoeval,chiang2024chatbot}.
%
%
While it is widely agreed that, given a sufficiently large set of pairwise comparisons, higher (lower) ranking under 
this paradigm corresponds to better (worse) human alignment, 
there have also been increasing concerns that this paradigm is too costly and time-consuming to be practical, 
especially given the pace at which models are updated and new models are developed.

To lower the cost and increase the efficiency of ranking from pairwise comparisons, 
it has become a common practice to ask a strong LLM---a model known to strongly align with human preferences---to perform pairwise comparisons~\cite{thomas2023large,wang2023large,chiang2023vicuna,chiang2023can,jiang2023llm,wang2024pandalm,qin2023ischatgpt,dubois2024alpacafarm,qin2023large,liu2024aligning}. 
The rationale is that, if a model strongly aligns with human pre\-fe\-ren\-ces, then, the distributions of pairwise comparisons by the model and by the human should in principle match~\cite{thomas2023large,chiang2023can,verma2023preference}.
Worrying\-ly, there are multiple lines of evidence,
including our experimental findings in Figure~\ref{fig:ranks-gpt4},
showing that the rankings constructed using pairwise comparisons made by a strong LLM are sometimes different to those constructed using pairwise comparisons by humans~\cite{zheng2023judging, li2023prd,boubdir2023elo,singhal2023large,dettmers2024qlora,davidson2024evaluating,hou2024large}, questioning the rationale above. 
In this work, we introduce a statistical framework to measure the uncertainty in the rankings constructed using pairwise comparisons made by a model, which may be introduced by a mismatch between human and model preferences or by the fact that we use a finite number of pairwise comparisons.

\xhdr{Our contributions}
Our framework measures uncertainty using rank-sets---sets of possible ranking positions that each model can take.
If the rank-set of a model is large (small), it means that there is high (low) uncertainty in the ranking position of the model. 
To construct the rank-sets, 
our framework first leverages a (small) set of pairwise comparisons by humans and a large set of pairwise comparisons by a strong LLM to create a confidence ellipsoid.
By using prediction-powered inference~\cite{angelopoulos2023prediction,angelopoulos2023ppi++,zrnic2024cross},
this confidence ellipsoid is guaranteed to contain the vector of (true) probabilities that each model is preferred over others by humans---the win-rates---with a user-spe\-ci\-fied coverage probability $1 - \alpha$.
Then, it uses the distance between this ellipsoid and the hyperplanes under which pairs of models have the same probability values of being preferred over others 
to efficiently cons\-truct the rank-sets.
Importantly, we can show that, with probability greater than or equal to $1-\alpha$, the constructed rank-sets are guaranteed to cover the ranking consistent with the (true) probability that each model is preferred over others by humans asymptotically.
Moreover, our framework does not make any assumptions on the distribution of human preferences nor about the degree of alignment between pairwise preferences of humans and the strong LLM.
Experiments on pairwise comparisons made by humans in the LMSYS Chatbot Arena platform~\cite{jiang2023llm} and pairwise comparisons made by three strong LLMs, namely GPT 3.5, Claude 3 and GPT 4, 
empirically demonstrate that the rank-sets constructed using our framework are 
more likely to cover the true ranking consistent with (the distribution of) human pairwise preferences than the rank-sets constructed using only pairwise comparisons made by the strong LLMs.
An open-source implementation of our methodology as well as the data on pairwise preferences of strong LLMs used in our experiments are available at \href{https://github.com/Networks-Learning/prediction-powered-ranking}{https://github.com/Networks-Learning/prediction-powered-ranking}.

\xhdr{Further related work}
Our work builds upon recent work on prediction-powered inference, ranking under uncertainty, and ran\-king of LLMs.

Prediction-powered inference~\cite{angelopoulos2023prediction,angelopoulos2023ppi++,zrnic2024cross} is a recently 
introduced statistical framework to obtain valid $p$-values and confidence intervals about a population-level quantity such as the mean outcome or a regression coefficient using a small labeled dataset and a large unlabeled dataset, whose labels are imputed using a black-box machine learning model.
However, our work is the first to use prediction-powered inference (as a subroutine) to construct rank-sets 
with coverage guarantees.
In this context, it is worth acknowledging that a very recent work by Saad-Falcon et al.~\cite{saadfalcon2023ares} has used prediction-powered inference to construct (single) rankings, rather than rank-sets. 
However, their rankings do not enjoy coverage guarantees with respect to the true ranking consistent with (the distribution of) the human preferences.
Moreover, an independent, concurrent work by Boyeau et al.~\cite{boyeau2024autoeval} has also used prediction-powered inference to construct (single) rankings based on the estimated coefficients of a Bradley-Terry model. 
However, the estimated coefficients come with large, overlapping confidence intervals, which would have led to
uninformative rank-sets, had the authors used them to construct rank-sets.

The vast majority of the literature on ranking under uncertainty has focused on confidence intervals for 
individual ranking positions~\cite{lemmers2007incorporating, goldstein1996league, hall2009using, marshall1998reliability, wright2014ranking, xie2009confidence, zhang2014confidence}. 
Only recently, a paucity of work has focused on joint measures of uncertainty for rankings~\cite{neuhof2023confident, rising2021uncertainty, al2022simultaneous, klein2020joint}.
Similarly as in our work, this line of work also seeks to construct rank-sets with coverage guarantees. 
However, 
in contrast to our work, 
it estimates the quality metric (in our work, the probability that an LLM is preferred over others) and the confidence intervals separately for each of the items (in our work, LLMs) using independent samples. 
As a consequence, it needs to perform multiple comparison 
correction to create the rank-sets.

In recent years, there has also been a flurry of work on ranking LLMs using benchmark datasets with manually hand-crafted inputs and ground-truth outputs~\cite{bach2022promptsource,wei2022finetuned,talmor2019commonsense,mishra2022cross,chen2021evaluating,liang2023holistic,longpre2023flan}.
However, it has become increasingly clear that oftentimes rankings derived from benchmark datasets do not 
correlate well with rankings derived from human preferences---an improved ranking position in the former
does not lead to an improved ranking position in the latter~\cite{zheng2023judging,li2023generative,li2023prd,chiang2023vicuna,chiang2024chatbot}.
Within the literature on ranking LLMs from pairwise comparisons, most studies use the Elo rating system~\cite{askell2021general,dettmers2024qlora,bai2022training,wu2023chatarena,lin2023llm}, originally introduced for chess tournaments~\cite{elo1966uscf}. 
However, Elo-based rankings are sensitive to the order of pairwise comparisons, as newer comparisons have more weight than older ones, which leads to unstable rankings~\cite{boubdir2023elo}.
To address this limitation, several studies have instead used the Bradley-Terry model~\cite{lmsys2023chatbot,chiang2024chatbot, boyeau2024autoeval}, which weighs pairwise comparisons equally regardless of their order. 
Nevertheless, both the Elo rating system and the Bradley-Terry model have faced criticism, 
as pairwise comparisons often fail to satisfy the fundamental axiom of transitivity, 
upon which both approaches rely~\cite{boubdir2023elo,bertrand2023limitations}, 
Recently, several studies have used the win-rate~\cite{zheng2023judging,chiang2024chatbot,boyeau2024autoeval}, which weighs comparisons equally regardless of their order and does not require the transitivity assumption, but requires humans to make pairwise comparisons between every pair of models.
In our work, we build upon the win-rate and lift the above requirement by using pairwise comparisons made by a strong LLM.

\section{LLM Ranking under Uncertainty}
\label{sec:preliminaries}
Let $\Mcal$ be a set of $k$ large language models (LLMs) and $P(Q)$ be a distribution of inputs on a discrete set 
of inputs $\Qcal$. 
Moreover, assume that, 
for each input $q \sim P(Q)$,\footnote{We denote random variables with capital 
letters and realizations of random variables with lower case letters.} each model $m \in \Mcal$ may provide an output 
$r \sim P_{m}(R \given Q=q)$ from a discrete set of outputs $\Rcal$.
Further, given two outputs $r, r' \in \Rcal$ from two different models,
the (binary) variables $w, w' \sim P(W, W' \given Q=q, R=r, R'=r')$  indicate whether a human prefers $r$ 
over $r'$ ($w=1$, $w'=0$) or viceversa ($w=0$, $w'=1$).
In the case of a tie, then $w=w'=0$.
In what follows, we use $m(r)$ and $m(r')$ to denote the models that provide outputs $r$ and $r'$ respectively,
and
without loss of generality, 
we assume that the output $r$ is shown first.
Then, our goal is to rank all models according to the (empirical) probability $\theta_m$ that their outputs are 
preferred over the outputs of any other model picked uniformly at random.

To this end, we start by writing the probability $\theta_m$ as an expectation over the distribution of inputs, outputs 
and pairwise preferences: 
\begin{equation}
\begin{split}
    \label{eq:theta-star}
    \theta_{m} = \frac{1}{k-1} \sum_{\tilde{m} \in \Mcal \backslash \{m\}} \EE_{Q} & \left[ \frac{1}{2} \EE_{R \sim P_m , R' \sim P_{\tilde{m}} }\left[ \EE_{W} [ W \given Q, R, R' ] \right] + \right. \\ & \left. \frac{1}{2} \EE_{R \sim P_{\tilde{m}}, R' \sim P_m }\left[\EE_{W'} [W' \given Q, R, R']\right]\right],    
\end{split}
\end{equation}
where note that the order of the pairs of outputs is picked at random.
Next, following previous work~\cite{al2022simultaneous,neuhof2023confident}, we formally cha\-rac\-te\-rize the ranking 
position of each model $m \in \Mcal$ in the ranking induced by the probabilities $\theta_m$ using a rank-set $[l(m), u(m)]$, 
where
\begin{equation}
    \label{eq:lower-upper-rank}
    l(m) = 1 + \sum_{\tilde{m} \in \Mcal\setminus \{m\}}\one\{\theta_{m} < \theta_{\tilde{m}} \} \quad \text{and} \quad
    u(m) = k - \sum_{\tilde{m} \in \Mcal\setminus \{m\}}\one\{\theta_{m} > \theta_{\tilde{m}} \}, 
\end{equation}
are the lower and upper ranking position respectively 
and smaller ranking position indicates better alignment with human preferences.
Here, note that it often holds that $\theta_{m} \neq \theta_{\tilde{m}}$ for all $\tilde{m} \in \Mcal\setminus \{m\}$ and then
the rank-set reduces to a singleton, \ie, $l(m) = u(m)$.

In general, we cannot directly construct the rank-sets as defined above because the probabilities $\theta_m$ are unknown. 
Consequently, the ty\-pi\-cal strategy reduces to first gathering pairwise comparisons by humans to compute unbiased 
estimates of the above probabilities using sample averages
and then construct estimates $[\hat{l}(m), \hat{u}(m)]$ of the rank-sets using Eq.~\ref{eq:lower-upper-rank} with $\hat{\theta}_m$ rather than $\theta_m$. 
Under this stra\-te\-gy, if the amount of pairwise comparisons we gather is sufficiently large, the estimates of the rank-sets will closely match the true rank-sets. 
However, since gathering pairwise comparisons from humans is costly and time-consuming, 
it has become a very common practice to gather pairwise comparisons $\hat{w}, \hat{w}'$ by a strong LLM, rather 
than pairwise comparisons $w, w'$ by humans~\cite{zheng2023judging,wang2024pandalm,liusie2023zero,li2023generative,dubois2024alpacafarm,jiang2023llm,li2023prd,li2023split,gao2023human, bai2024benchmarking},
and then utilize them to compute unbiased estimates of the probabilities $\breve{\theta}_m$ that the outputs provided by 
each model is preferred over others by the strong LLM, 
which can be written in terms of expectations as follows:
\begin{equation}
\begin{split}
    \label{eq:theta-model}
    \breve{\theta}_{m} = \frac{1}{k-1} \sum_{\tilde{m} \in \Mcal \backslash \{m\}} \EE_{Q} & \left[ \frac{1}{2} \EE_{R \sim P_m , R' \sim P_{\tilde{m}} }\left[ \EE_{\hat{W}}[\hat{W} \given Q, R, R'] \right] + \right. \\ & \left. \frac{1}{2} \EE_{R \sim P_{\tilde{m}}, R' \sim P_m }\left[\EE_{\hat{W}'}[\hat{W}' \given Q, R, R'] \right]\right].    
\end{split}
\end{equation}
In general, one can only draw valid conclusions about $\thetab$ using (an estimate of) $\breve{\thetab}$ if the distribution 
of the pairwise comparisons by the strong LLM $P(\hat{W}, \hat{W}' \given Q=q, R=r, R'=r')$ closely matches the 
distribution of pairwise comparisons by the humans $P(W, W' \given Q=q, R=r, R'=r')$ for any $q \in \Qcal$ and $r, r' \in \Rcal$.
However, 
there are multiple lines of evidence showing that there is a mismatch between the distributions,
questioning the validity of the conclusions drawn by a myriad of papers.
In what follows, we introduce a statistical framework that, 
by complementing a (large) set of $N+n$ pairwise comparisons $\hat{w}, \hat{w}'$ by a strong LLM with a small set of $n$ pairwise comparisons $w, w'$ by humans, 
is able to construct estimates $[\hat{l}(m), \hat{u}(m)]$ of the rank-sets with provable co\-ve\-rage guarantees.
More formally, given a user-specified value $\alpha \in (0, 1)$, the estimates of the rank-sets 
satisfy that
\begin{equation}
\label{eq:confident-rank-sets}
    \lim_{n} \PP \left( \bigcap_{m \in \Mcal} [l(m), u(m)] \subseteq [\hat{l}(m), \hat{u}(m) ] \right) \geq 1 - \alpha.
\end{equation}
To this end, we will first use prediction-powered inference~\cite{angelopoulos2023prediction,angelopoulos2023ppi++} to construct a confidence ellipsoid that, with probability $1 - \alpha$, is gua\-ran\-teed to contain the (column) vector of (true)
probabilities $\thetab = (\theta_m)_{m \in \Mcal}$.
Then, we will use the distance between this ellipsoid and the hyperplanes under which each pair 
of models $m, \tilde{m} \in \Mcal$ have the same probability values of being preferred over others, 
to efficiently cons\-truct the estimates $[\hat{l}(m), \hat{u}(m)]$ of the rank-sets. 

\section{Constructing Confidence Regions with Prediction-Powered Inference}
\label{sec:theta}
Let the set $\Dcal_N = \{ (q_i, r_i, r'_i, m(r_i), m(r_i'), \hat{w}_i, \hat{w}'_i) \}_{i=1}^{N}$ comprise pairwise comparisons by a strong LLM to $N$ inputs and the set $\Dcal_n = \{ (q_i, r_i, r'_i, m(r_i), m(r'_i), w_i, w'_i, \hat{w}_i, \hat{w}'_i) \}_{i=1}^{n}$ comprise pairwise comparisons by the same strong LLM and by humans to $n$ inputs, with $n \ll N$.
In what follows, for each pairwise comparison, we will refer to the models $m(r)$ and $m(r')$ that provided the first and second output using one-hot (column) vectors $\mb$ and $\mb'$, respectively. 
Moreover, to summarize the pairwise comparisons\footnote{We assume that each model $m \in \Mcal$ participates 
in at least one pairwise comparison in both $\Dcal_N$ and $\Dcal_n$.} in $\Dcal_{N}$ and $\Dcal_{n}$,
we will stack the one-hot vectors $\mb$ and $\mb'$ into four matrices, $\Mb_N$ and $\Mb'_{N}$ for $\Dcal_{N}$ and $\Mb_n$ and $\Mb'_n$ for $\Dcal_n$, where each column corresponds to a one-hot vector, and the indicators $w$ and $\hat{w}$ into six (column) vectors, $\hat{\wb}_{N}$ and $\hat{\wb}'_{N}$ for $\Dcal_{N}$ and $\hat{\wb}_n$, $\hat{\wb}'_{n}$, $\wb_n$ and $\wb'_n$ for $\Dcal_{n}$.

\begin{algorithm}[ht]
\DontPrintSemicolon
\small
\caption{It estimates $\hat{\thetab}$ and $\widehat{\Sigmab}$ using prediction-powered inference.}
\label{alg:1}
\KwIn{$k$, $\Dcal_{N}$, $\Dcal_n$}
\KwOut{$\hat{\thetab},\widehat{\Sigmab}$}

\vspace{2mm}
$\hat{\wb}_N, \hat{\wb}'_N,\Mb_N,\Mb'_N \gets$ \textsc{summarize}$(\Dcal_N,k)$

$\wb_n,\wb'_n,\hat{\wb}_n,\hat{\wb}'_n,\Mb_n,\Mb'_n \gets$\textsc{summarize}$(\Dcal_n,k)$

\vspace{2mm}
$\lambda \gets$ \textsc{lambda}$(k,\hat{\wb}_N, \hat{\wb}'_N,\Mb_N,\Mb'_N,\wb_n,\wb'_n,\hat{\wb}_n,\hat{\wb}'_n,\Mb_n,\Mb'_n)$ \tcp*{Algorithm~\ref{alg:lambda}}

$\ab \gets \left(  \one_{k} \left(\left(\Mb_{N}+\Mb'_{N}\right)\one_{N}\right)^{\top}\odot \II_k \right)^{-1} \left( \Mb_N \cdot \lambda\hat{\wb}_N + \Mb'_N \cdot\lambda\hat{\wb}'_N \right)$

$\bb \gets \left(  \one_{k} \left(\left(\Mb_{n}+\Mb'_{n}\right)\one_{n}\right)^{\top}\odot \II_k\right)^{-1}\left( \Mb_n (\lambda \hat{\wb}_n - \wb_n) + \Mb'_n\left( \lambda\hat{\wb}'_n-\wb'_n \right) \right)$

$\hat{\thetab}\gets \ab-\bb$ \tcp*{prediction-powered estimate (Eq.~\ref{eq:theta-hat})}

\vspace{2mm}

$\Ab \gets \left( \left( \one_{k}(\lambda\hat{\wb}_{N}-\Mb^{\top}_{N}\ab)^{\top} \right)\odot \Mb_{N} + \left( \one_{k}(\lambda\hat{\wb}'_{N}-\Mb'^{\top}_{N}\ab)^{\top} \right)\odot \Mb'_{N} \right)$

$\Bb \gets \left(\one_{k}(\lambda\hat{\wb}_{n}-\wb_n-\Mb^{\top}_{n}\bb)^{\top} \right)\odot \Mb_{n} + \left( \one_{k}(\lambda\hat{\wb}'_{n}-\wb'_n-\Mb'^{\top}_{n}\bb)^{\top} \right)\odot \Mb'_{n}$

$\widehat{\Sigmab} \gets \frac{1}{N^2} \Ab \Ab^{\top} + \frac{1}{n^2} \Bb \Bb^{\top}$\,\tcp*{estimate of covariance (Eq.~\ref{eq:sigma})}

\Return{$\hat{\thetab},\widehat{\Sigmab}$} 
\end{algorithm}

Then, building upon the recent framework of prediction-powered inference~\cite{angelopoulos2023prediction}, we compute an unbiased estimate $\hat{\thetab}$ of the vector of (true) probabilities $\thetab$:
\begin{multline}
\label{eq:theta-hat}
    \hat{\thetab} = \underbrace{ \left(   \one_{k} \left(\left(\Mb_{N}+\Mb'_{N}\right)\one_{N}\right)^{\top}\odot \II_k\right)^{-1} \left( \Mb_N \cdot \lambda\hat{\wb}_N + \Mb'_N \cdot\lambda\hat{\wb}'_N \right) }_{\ab} \\
    - \underbrace{ \, \left(  \one_{k} \left(\left(\Mb_{n}+\Mb'_{n}\right)\one_{n}\right)^{\top}\odot \II_k\right)^{-1} \left( \Mb_n (\lambda \hat{\wb}_n - \wb_n) + \Mb'_n\left( \lambda\hat{\wb}'_n-\wb'_n \right) \right) }_{\bb},
\end{multline}
where 
$\one_d$ denotes a $d$-dimensional column vector where each dimension has value $1$ 
and $\II_k$ denotes a $k$-dimensional identity matrix.
Here, note that the first term $\ab$ utilizes the pairwise comparisons by the strong LLM from $\Dcal_{N}$ to compute
an unbiased estimate of the vector of probabilities $\breve{\thetab}$ defined in Eq.~\ref{eq:theta-model} using sample
averages,
and the second term $\bb$ utilizes the pairwise comparisons by the strong LLM and by humans from $\Dcal_{n}$ to compute an unbiased estimate of the difference of probabilities $\thetab - \breve{\thetab}$ defined in Eqs.~\ref{eq:theta-star} and~\ref{eq:theta-model}, also using sample averages. The parameter $\lambda \in [0,1]$ weighs the comparisons $\hat{\wb},\hat{\wb}'$ differently than the comparisons $\wb, \wb'$.
Details on why this can be useful and on the selection of $\lambda$ are in Appendix~\ref{app:lambda}.

Further, as shown in Angelopoulos et al.~\cite{angelopoulos2023ppi++}, 
the difference of probabilities $\hat{\thetab} - \thetab$ converges in distribution to a $k$-dimensional normal 
$\Ncal_{k}(0, \Sigmab)$, where $\Sigmab = \EE[ (\hat{\thetab} - \thetab) (\hat{\thetab} - \thetab)^{\top} ]$, 
and thus the confidence region
\begin{equation}
\label{eq:ellipsoid}
    \Ccal_{\alpha}=\left \{\xb \in \RR^k \given 
    \left(\xb-\hat{\thetab}\right)^{\top}\left( \frac{\widehat{\Sigmab}^{-1}}{ \chi^2_{k,1-\alpha}} \right) \left(\xb-\hat{\thetab} \right) \leq 1 \right \},%
\end{equation}
where $\widehat{\Sigmab}$ is an empirical estimate of the covariance matrix $\Sigmab$ using pairwise comparisons from $\Dcal_N$ 
and $\Dcal_n$, \ie, 
\begin{equation} \label{eq:sigma}
\widehat{\Sigmab} = \frac{1}{N^2} \Ab \Ab^{\top} + \frac{1}{n^2} \Bb \Bb^{\top},
\end{equation}
with 
\begin{align*}
\Ab &= \left( \left( \one_{k}(\lambda\hat{\wb}_{N}-\Mb^{\top}_{N}\ab)^{\top} \right)\odot \Mb_{N} + \left( \one_{k}(\lambda\hat{\wb}'_{N}-\Mb'^{\top}_{N}\ab)^{\top} \right)\odot \Mb'_{N} \right), \\
\Bb &= \left(\one_{k}(\lambda\hat{\wb}_{n}-\wb_n-\Mb^{\top}_{n}\bb)^{\top} \right)\odot \Mb_{n} + \left( \one_{k}(\lambda\hat{\wb}'_{n}-\wb'_n-\Mb'^{\top}_{n}\bb)^{\top} \right)\odot \Mb'_{n},
\end{align*}
and $\chi^2_{k,1-\alpha}$ is the $1 - \alpha$ quantile of the $\chi^2$ distribution with $k$ degrees of freedom, 
satisfies that
\begin{equation} \label{eq:1-alpha-guarantee}
\lim_{n} \PP(\thetab \in \Ccal_{\alpha}) = 1 - \alpha.
\end{equation}
Algorithm~\ref{alg:1} summarizes the overall procedure to compute $\hat{\thetab}$ and $\widehat{\Sigmab}$, which runs in $O(k^2(N+n))$ time.

\section{Constructing Rank-Sets with Coverage Guarantees}
\label{sec:rank-sets}
\begin{algorithm}[ht]
\DontPrintSemicolon
\small
\caption{It constructs $[\lhat(m),\hat{u}(m)]$ for all $m \in \Mcal$}
\label{alg:2} 
\KwIn{$\Mcal$, $\Dcal_N$, $\Dcal_n$, $\alpha$}
\KwOut{$\{[\lhat(m),\hat{u}(m)]\}_{m \in \Mcal}$}

$k \gets |\Mcal|$\;
$\hat{\thetab}, \widehat{\Sigmab} \gets$ \textsc{confidence-ellipsoid}$(k,\Dcal_N,\Dcal_n)$ \tcp*{Algorithm~\ref{alg:1}}
\vspace{1mm}
\For{$m\in\Mcal$} {
    $\lhat(m) \gets 1$, \quad $\hat{u}(m) \gets k$\;
    \For{$\tilde{m} \in \Mcal \setminus \{m\}$} {
        {$d \gets \frac{|\thetahat_{m} - \thetahat_{\tilde{m}}| }{\sqrt{2}} - \sqrt{ \frac{1}{2}(\widehat{\Sigma}_{m,m}+\widehat{\Sigma}_{\tilde{m},\tilde{m}}-2\widehat{\Sigma}_{m,\tilde{m}})\chi^{2}_{k, 1 - \alpha}}$} 
        \tcp*{Eq.~\ref{eq:distance-ellipsoid-hyperplane}}
        \If{$d>0$ \textbf{and} $\thetahat_m<\thetahat_{\tilde{m}}$} {
            $\lhat(m) \gets \lhat(m) +1$\;
        }
        \ElseIf{$d>0$ \textbf{and} $\thetahat_m>\thetahat_{\tilde{m}}$}{
            $\hat{u}(m) \gets \hat{u}(m)-1$\;
        }
    }
}
\Return{$\{[\lhat(m),\hat{u}(m)]\}_{m \in \Mcal}$}
\end{algorithm}

For each pair of models $m, \tilde{m} \in \Mcal$ such that $m \neq \tilde{m}$, we first define a hyperplane $H_{m, \tilde{m}} \subseteq \RR^ {k}$ 
as follows: 
\begin{equation}
\label{eq:hyperplane}
    H_{m, \tilde{m}} = \{\xb \in \RR^{k}\given x_m = x_{\tilde{m}}\}.
\end{equation}
Then, for each of these hyperplanes $H_{m, \tilde{m}}$, we calculate the distance $d(\Ccal_{\alpha}, H_{m,\tilde{m}})$ 
between $H_{m, \tilde{m}}$ and the confidence region $\Ccal_{\alpha}$ defined by Eq.~\ref{eq:ellipsoid}, \ie,
\begin{equation}
    \label{eq:distance-ellipsoid-hyperplane}
    d(\Ccal_{\alpha}, H_{m,\tilde{m}}) = \frac{|\thetahat_m-\thetahat_{\tilde{m}}|
    -\sqrt{(\widehat{\Sigma}_{m,m}+\widehat{\Sigma}_{\tilde{m},\tilde{m}}-2\widehat{\Sigma}_{m,\tilde{m}})\chi^2_{k,1-\alpha}}}
    {\sqrt{2}},
\end{equation}
where $\widehat{\Sigmab}$ is the empirical covariance matrix defined by Eq.~\ref{eq:sigma}.

Now, for each pair of models $m, m' \in \Mcal$, we can readily conclude that, 
if the distance $d(\Ccal_{\alpha}, H_{m,\tilde{m}}) > 0$, 
then, 
the confidence region $\Ccal_{\alpha}$ either lies in the half-space of $\RR^{k}$ where $x_m > x_{\tilde{m}}$ 
if $\hat{\theta}_m > \hat{\theta}_{\tilde{m}}$
%
or it lies in the half space of $\RR^{k}$ where $x_{m} < x_{\tilde{m}}$ if 
$\hat{\theta}_m < \hat{\theta}_{\tilde{m}}$.
Building upon this observation, for each model $m \in \Mcal$, we construct the following 
estimates $[\hat{l}(m), \hat{u}(m)]$ of the rank-sets $[l(m), u(m)]$:
\begin{equation} \label{eq:estimates-rank-sets}
\begin{split}
    \hat{l}(m) &= 1 + \sum_{\tilde{m} \in \Mcal\setminus\{m\}}\one\{ d(\Ccal_{\alpha}, H_{m,\tilde{m}}) > 0\} \cdot \one\{ \thetahat_{m} < \thetahat_{\tilde{m}}\}\\
    \hat{u}(m) &= k -  \sum_{\tilde{m} \in \Mcal\setminus\{m\}}\one\{ d(\Ccal_{\alpha}, H_{m,\tilde{m}}) > 0\} \cdot \one\{ \thetahat_{m} > \thetahat_{\tilde{m}}\}.
\end{split}
\end{equation}
Importantly, 
using a similar proof technique as in Lemma~1 in Neuhof and Benjamini~\cite{neuhof2023confident},
we can show that the above rank-sets estimates 
enjoy provable co\-ve\-rage guarantees with respect to the rank-sets $[l(m), u(m)]$ 
induced by the probabilities $\thetab$ 
that the outputs of each model is preferred over any other model by humans (proven in Appendix~\ref{app:proof}): %
\begin{theorem}
\label{theorem:coverage}
The estimates $[\hat{l}(m), \hat{u}(m)]$ of the rank-sets defined by Eq.~\ref{eq:estimates-rank-sets}
satisfy that
\begin{equation} \label{eq:estimates-rank-sets-coverage}
    \lim_{n}\PP \left ( \bigcap_{m \in \Mcal}[l(m), u(m)] \subseteq [\hat{l}(m), \hat{u}(m)] \right) \geq 1 - \alpha.
\end{equation}
\end{theorem}
\noindent Algorithm~\ref{alg:2} summarizes the overall procedure to construct the rank-sets $[\hat{l}(m), \hat{u}(m)]$ for all $m \in \Mcal$, which runs in $O(k^2(N+n))$.

\section{Experiments}
\label{sec:experiments}
We apply our framework to construct rank-sets for $12$ popular LLMs  using
pairwise comparisons made by humans in the LMSYS Chatbot Arena platform\footnote{\href{https://chat.lmsys.org/}{https://chat.lmsys.org/}} 
and pairwise comparisons made by three strong LLMs. 
We show that the rank-sets constructed using our framework are significantly more likely to cover the true ranking consistent with (the distribution of) human pairwise
preferences than the rank-sets constructed using only pairwise comparisons made by the
strong LLMs. 

\xhdr{Experimental setup} Our starting point is the Chatbot Arena dataset~\cite{zheng2023judging}, which comprises $33{,}481$ pairwise comparisons made by $13{,}383$ humans about the responses given by $20$ different LLMs to $26{,}968$ unique queries.
In what follows, we refer to each pair of responses to a query by two different LLMs and the query itself as an instance.
As an initial pre-processing, we filter out any instance whose corresponding query is flagged as toxic or multiturn.
Then, we gather pairwise comparisons made by three strong LLMs, namely GPT-$3.5$-turbo-$0125$ (\gptthree), GPT-$4$-$0125$-preview (\gptfour) and Claude-$3$-Opus-$20240229$ (\claude), about all the (pre-processed) instances from the Chatbot Arena dataset.
To this end, we use (almost) the same prompt as in Zheng et al.~\cite{zheng2023judging},
which instructs each strong LLM to output option `A' (`B') if it prefers the response of first (second) LLM, or option `C' if it declares a tie. 
Further, we filter out any instances for which at least one strong LLM provides a verbose output instead of `A', `B', or `C', and focus on 
a set of LLMs
with at least $96$ pairwise comparisons between every pair of LLMs in the set.
After these pre-processing steps, we have $14{,}947$ instances comprising $13{,}697$ unique queries and $12$ different LLMs and, for each instance, we have one pairwise comparison made by a human and three pairwise comparisons by the three strong LLMs.
Refer to Appendix~\ref{app:experimental-setup} for more information regarding the $12$ LLMs, the number of pairwise comparisons between every pair of LLMs, and the prompt used to gather pairwise comparisons made by the three strong LLMs.

To draw reliable conclusions, in each experiment, we construct rank-sets $1{,}000$ times and, each time, we use a random set of $N + n = 6{,}336$ instances with an equal number of instances per pair of models, out of the $14{,}947$ instances.
The values of $N$ and $n$ vary across experiments and they define two random subsets, also with an equal number of instances per pair of models.

\xhdr{Methods} In our experiments, we construct rank-sets using the following methods:
\squishlist
\item[a)] \baseline: it constructs (unbiased) rank-sets using the pairwise comparisons made by humans corresponding to the random set of $N+n$ instances via Algorithms~\ref{alg:2} and~\ref{alg:1b}, shown in Appendix~\ref{app:alg-conf}. 
The constructed rank-sets are presumably likely to cover the true rank-sets.

\item[b)] \llm~\gptfour, \llm~\gptthree and \llm~\claude:
they construct (possibly biased) rank-sets using the pairwise comparisons made by one of the three strong LLMs corresponding to the random set of $N+n$ instances via Algorithms~\ref{alg:2} and~\ref{alg:1b}.

\item[c)] \ppr~\gptfour, \ppr~\gptthree and \ppr~\claude:
they construct (unbiased) rank-sets using pairwise comparisons made by one of the three strong LLMs corresponding to the random set of $N+n$ instances and pairwise comparisons made by humans corresponding to the random subset of $n$ instances via Algorithms~\ref{alg:1} and~\ref{alg:2}.

\item[d)] \human: it constructs (unbiased) rank-sets using the pairwise comparisons made by humans corresponding to the random subset of $n$ instances via Algorithms~\ref{alg:2} and~\ref{alg:1b}.
\squishend
In the above, note that a), b) and d) use linear regression to construct a confidence region $\Ccal_{\alpha}$ using only pairwise comparisons by either humans or a strong LLM via Algorithm~\ref{alg:1b}, which runs in $O(k^2 (N+n))$ in a)-b) and in $O(k^2 n)$ in d), and then use this confidence region to construct 
the rank-sets via Algorithm~\ref{alg:2}.

\xhdr{Quality metrics}
Since the true probabilities $\thetab$ are unknown, we cannot compute the true rank-sets of the $12$ LLMs under comparison, which presumably may be singletons. 
As a result, we cannot estimate the (empirical) coverage probability---the probability that the rank-sets constructed using the above methods cover the true rank-sets, which Theorem~\ref{theorem:coverage} refers to.
To overcome this, we assess the quality of the rank-sets using two alternative metrics: 
rank-set size and baseline intersection probability.
Here, smaller (larger) rank-set sizes and larger (smaller) intersection probabilities are better (worse).
The baseline intersection probability is just the (empirical) probability that the rank-sets $[\hat{l}(m),\hat{u}(m)]$ constructed using one of the above methods intersect with the rank-sets $[\tilde{l}(m),\tilde{u}(m)]$ constructed using the \baseline method, \ie, 
%
$\PP \left( \bigcap_{m\in\Mcal}\one \left\{[\tilde{l}(m),\tilde{u}(m)] \cap [\hat{l}(m),\hat{u}(m)]\right\} \right)$.
%
Intuitively, we expect
that the larger the baseline intersection probability, the larger the coverage probability since the \baseline method uses a large(r) number of pairwise comparisons by humans to construct (unbiased) rank-sets and thus it is expected to approximate well the true rank-sets.
%
Further, note that the baseline intersection probability tells us how frequently there exists at least one single ranking covered by both one of the above methods and the \baseline method.
In Appendix~\ref{app:baseline-coverage-probability}, we experiment with an alternative metric, namely baseline coverage probability, which is the (empirical) probability that the rank-sets constructed using one of the above methods covers the rank-sets constructed using the \baseline method.
In Appendix~\ref{app:synthetic}, we additionally evaluate our framework in a synthetic setting where the true rank-sets are known, allowing us to compute the coverage probability and rank-biased overlap (RBO)~\cite{webber2010similarity}.

\begin{figure}
\includegraphics[width=0.95\textwidth]{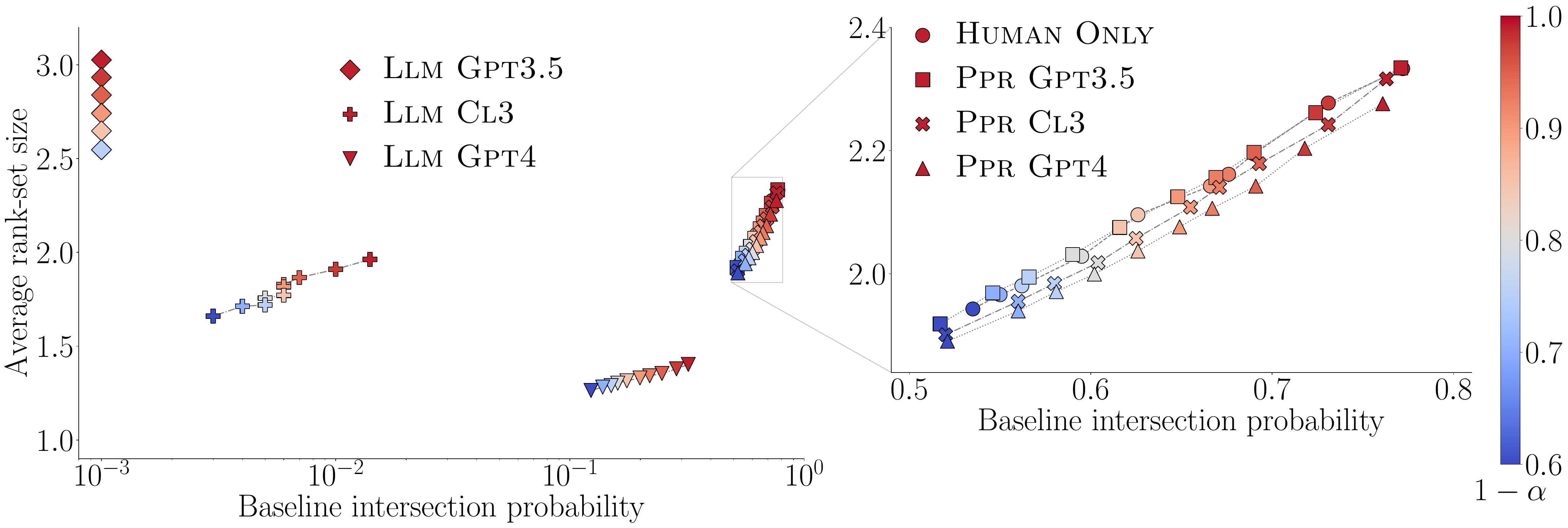}
\caption{Average rank-set size against baseline intersection probability for rank-sets constructed using only pairwise comparisons by a strong LLM (\llm~\gptfour, \llm~\gptthree and \llm~\claude), 
only pairwise comparisons by humans (\human), 
and pairwise comparisons by both a strong LLM and humans (\ppr~\gptfour, \ppr~\gptthree and \ppr~\claude) for different values of $\alpha$ and $n=990$. 
Smaller (larger) average rank-set sizes and larger (smaller) intersection probabilities are better (worse). In all panels, $95\%$ confidence bars for the rank-set size are not shown, as they are always below $0.02$.
} \label{fig:rankset-size-vs-intersection-rate}
\vspace{-1mm}
\end{figure}

\begin{figure}[t]
\includegraphics[width=1\textwidth]{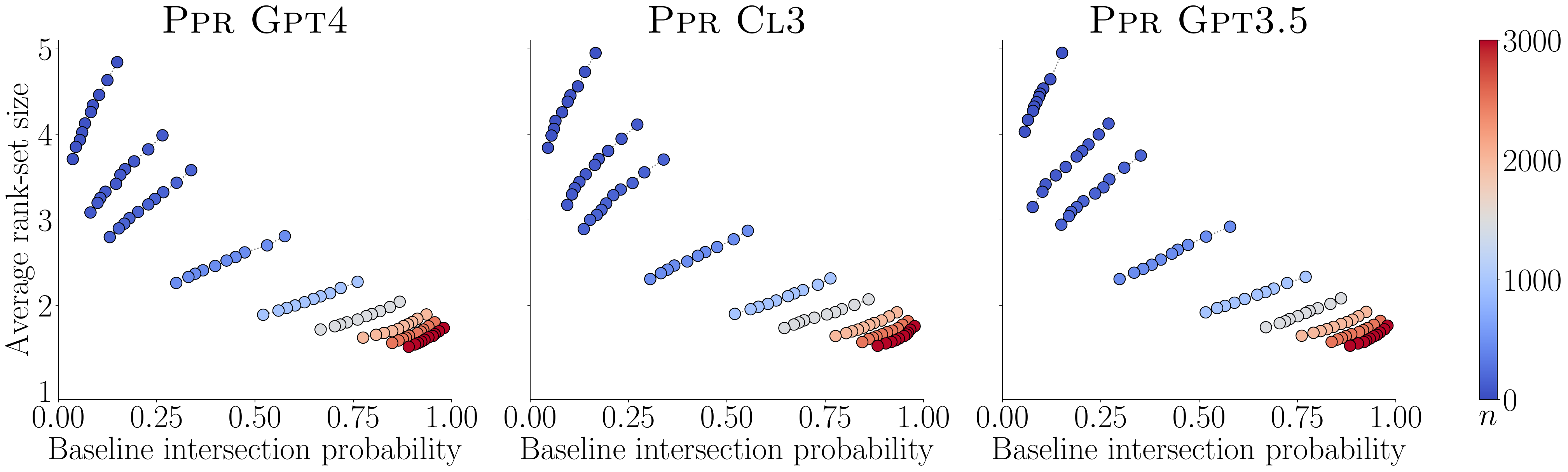}
\caption{Average rank-set size against baseline intersection probability for rank-sets constructed using pairwise comparisons by both a strong LLM and humans for different values of $n$ and $\alpha$. Smaller (larger) average rank-set sizes and larger (smaller) intersection probabilities are better (worse). In all panels, $95\%$ confidence bars for the rank-set size are not shown, as they are always below $0.04$.} 
\label{fig:vary-n-ppr}
\vspace{-2mm}
\end{figure}

\newpage
\xhdr{Quality of the rank-sets}
Figure~\ref{fig:rankset-size-vs-intersection-rate} shows the average rank-set size against the baseline intersection probability for
rank-sets constructed using all methods\footnote{In Appendix~\ref{app:confidence-intervals-intersection}, we include a version of this figure with three panels, where each panel contains the results for one strong LLM and confidence regions for the rank-set sizes.} except \baseline for different $\alpha$ values\footnote{$\alpha \in \{0.4, 0.3, 0.25, 0.2, 0.15, 0.1, 0.075, 0.05, 0.025, 0.01\}$.} and $n = 990$. 
The results show several interesting insights.
First, we find that rank-sets constructed using only pairwise comparisons by a strong LLM (\llm~\gptfour, \llm~\gptthree and \llm~\claude) achieve much lower baseline intersection probability, even orders of magnitude lower, than those constructed using only pairwise comparisons by humans (\human) or using both pairwise comparisons by a strong LLM and humans (\ppr~\gptfour, \ppr~\gptthree and \ppr~\claude).
This suggests that the distributions of pairwise comparisons by strong LLMs and humans are actually different, questioning the rationale used by an extensive line of work that proposed using only pairwise comparisons by strong LLMs to rank LLMs~\cite{wang2023large,chiang2023vicuna,chiang2023can,jiang2023llm,wang2024pandalm,dubois2024alpacafarm,zheng2023judging}.
Second, we find that rank-sets constructed using both pairwise comparisons by two of the strong LLMs and humans (\ppr~\gptfour and \ppr~\claude) achieve a better
trade-off between average rank-set size and baseline intersection probability than those constructed using only pairwise comparisons by humans (\human).
This suggests that pairwise comparisons by strong LLMs are valuable if they are complemented with (a few) pairwise comparisons by humans.
Third, we find that, among the three strong LLMs, GPT 4 stands out as the best performer.

Figure~\ref{fig:vary-n-ppr} shows the average rank-set size against the baseline intersection probability for rank-sets constructed using \ppr~\gptfour, \ppr~\gptthree and \ppr~\claude 
for different values of $n$ and $\alpha$ (the same values as in Figure~\ref{fig:rankset-size-vs-intersection-rate}).\footnote{$n \in \{66, 132, 198, 462, 990, 1452, 1980, 2442, 2970\}$.}
The results show that the trade-off between rank-sets and baseline intersection probabilities improves rapidly as the number of pairwise comparisons by humans $n$ increases but with diminishing returns.

\begin{figure}[t]
\includegraphics[width=\textwidth]{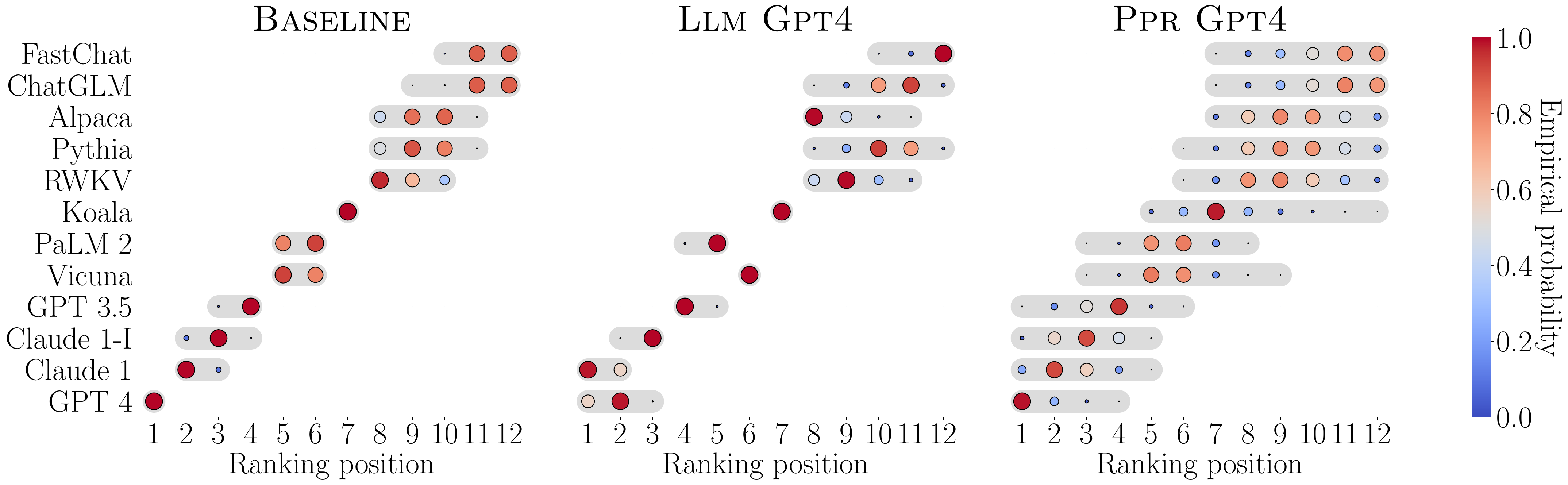}
\caption{Empirical probability that each ranking position is included in the rank-sets constructed by \baseline, \llm~\gptfour and \ppr~\gptfour for each of the LLMs under comparison. In all panels, $n=990$ and $\alpha=0.05$.
Larger (smaller) dots indicate higher (lower) empirical probability.} \label{fig:ranks-gpt4}
\vspace{-2mm}
\end{figure}

\begin{figure}[t]
\includegraphics[width=\textwidth]{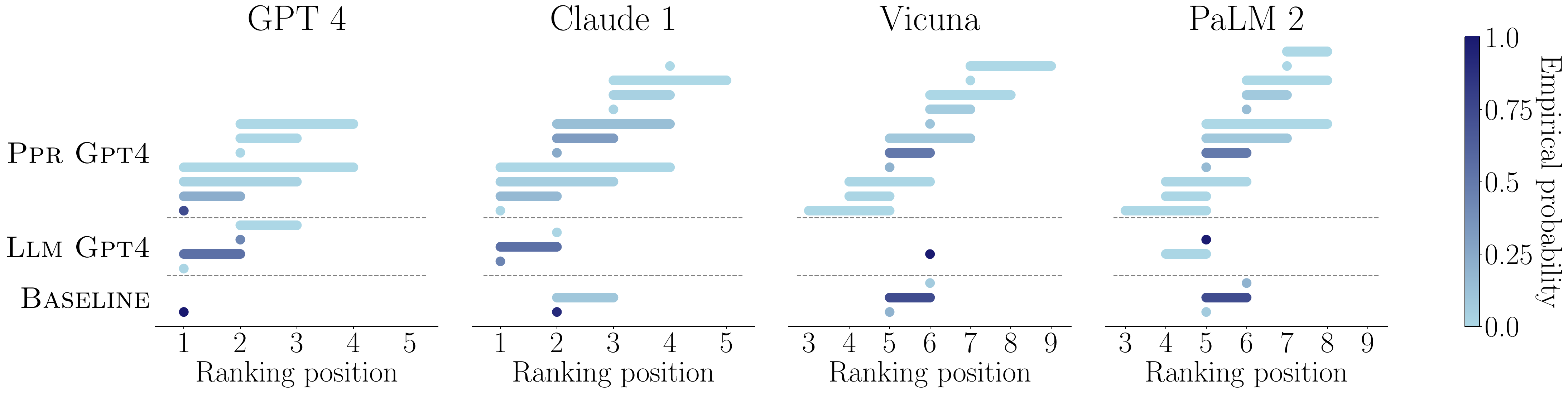}
\caption{Empirical probability of each rank-set constructed by \baseline, \llm~\gptfour and \ppr~\gptfour for GPT 4 (left), Claude 1 (middle left), Vicuna (middle right) and PaLM 2 (right). In all panels, $n=990$ and $\alpha=0.05$.} \label{fig:ranksets-claude1}
\vspace{-1mm}
\end{figure}

\xhdr{Structure of the rank-sets} 
In this section, we take a closer look to the structure of the rank-sets constructed using \baseline, \llm~\gptfour and \ppr~\gptfour. In Appendix~\ref{app:structure-of-ranksets}, we include additional results for all other methods.

First, we compute the empirical probability that each ranking position is included in the rank-sets constructed by \baseline, \llm~\gptfour and \ppr~\gptfour of each of the LLMs under comparison. 
Figure~\ref{fig:ranks-gpt4} summarizes the results for $n=990$ and $\alpha=0.05$, which reveal several interesting insights.
We find that there is lower uncertainty regarding the ranking position of each model for \llm~\gptfour than for \ppr~\gptfour. 
However, for \llm~\gptfour, the ranking position with the highest probability mass differs from \baseline in $7$ out of $12$ LLMs, 
including the top-2 performers. 
In contrast, for \ppr~\gptfour, it only differs from \baseline in $3$ out of $12$ LLMs.
This questions once more the status quo, which proposed using only pairwise comparisons by strong LLMs to rank LLMs~\cite{wang2023large,chiang2023vicuna,chiang2023can,jiang2023llm,wang2024pandalm,dubois2024alpacafarm,zheng2023judging}.

Next, we compute the empirical probability of each rank-set constructed by \baseline, \llm~\gptfour and \ppr~\gptfour for each of the LLMs under comparison.
Figure~\ref{fig:ranksets-claude1} summarizes the results for GPT 4, Claude 1, Vicuna and PaLM 2 for $n=990$ and $\alpha=0.05$. 
In agreement with the findings derived from Figure~\ref{fig:ranks-gpt4}, we observe that the distribution of rank-sets constructed by \llm~\gptfour is more concentrated than the distribution of rank sets constructed by \ppr~\gptfour. 
However, the rank-sets with the highest probability mass constructed by \llm~\gptfour coincide with those constructed by \baseline much less frequently than those constructed by \ppr~\gptfour.
Refer to Appendix~\ref{app:structure-of-ranksets} for qualitatively similar results for other LLMs.

\section{Discussion and Limitations}
\label{sec:discussion}
In this section, we highlight several limitations of our work, discuss its broader impact, and propose avenues for future work.

\xhdr{Data}
Our framework assumes that the queries and the pairwise comparisons made by humans and the strong LLMs are drawn i.i.d. from fixed distributions.
In future work, it would be very interesting to lift these assumptions and allow for distribution shift.   
Moreover, our framework assumes that the pairwise comparisons made by humans are truthful. 
However, an adversary could have an economic incentive to make pairwise comparisons strategically in order to favor a specific model over others. 
In this context, it would be interesting to extend our framework so that it is robust to strategic behavior.

\xhdr{Methodology}
Our framework utilizes rank-sets as a measure of uncertainty in rankings. However, in case of limited pairwise comparison data, rank-sets may be large and overlapping, reducing their value. 
In such situations, it may be worthwhile to explore other measures of uncertainty for rankings beyond rank-sets.
Further, to measure the level of alignment with human preferences, our framework utilizes the win-rate---the probability that the outputs of each model are preferred over the outputs of any other model picked uniformly at random.
However, if we need to rank $k$ LLMs and $k$ is \emph{large}, win-rate may be impractical since, to obtain reliable estimates, we need to gather $O(k^2)$ pairwise comparisons made by humans.
Finally, our framework
constructs
rank-sets with asymptotic coverage guarantees, however, it would be interesting to derive PAC-style, finite-sample coverage guarantees.

\xhdr{Evaluation}
We have showcased our framework using pairwise comparisons made by humans in a single platform, namely LMSYS Chatbot Arena, and pairwise comparisons made by just three strong LLMs. 
As a result, one may question the generalizability of the conclusion derived from the rank-sets estimated using our framework.
In this context, it is also important to acknowledge that, in LMSYS Chatbot Arena, the queries are chosen by the humans who make pairwise comparisons and this may introduce a variety of biases.
Therefore, it would be interesting to apply our framework to human data from other platforms.

\xhdr{Broader Impact}
Our framework rank LLMs according to their level of alignment with human preferences---a LLM is ranked higher than others if its outputs are more frequently preferred by humans. However, in many application domains, especially in high-stakes scenarios, it may be important to account for other important factors such as accuracy, fairness, bias and toxicity~\cite{liang2023holistic}.

\section{Conclusions}
\label{sec:conclusions}
We have introduced a statistical framework to construct a ranking of a collection of LLMs consistent with their level of alignment with human preferences 
using a small set of pairwise comparisons by humans and a large set of pairwise comparisons by a strong LLM.
Our framework quantifies uncertainty in the ranking by providing a rank-set---a set of possible ranking positions---for each of the models under comparison. 
Moreover, it guarantees that, with a probability greater than or equal to a user-specific value, the rank-sets cover the ranking consistent with the (true) probability that each model is preferred over others by humans asymptotically.
Finally, we have empirically demonstrated that the rank-sets constructed using our framework are more likely to cover the true ranking consistent with (the distribution of) human pairwise preferences than the rank-sets constructed using only pairwise comparisons made by the strong LLMs.

\vspace{1mm}
\xhdr{Acknowledgements} Gomez-Rodriguez acknowledges support from the European Research Council (ERC) under the European Union'{}s Horizon 2020 research and innovation programme (grant agreement No. 945719).

\newpage

\bibliographystyle{unsrt}
\bibliography{refs}

\clearpage
\newpage

\appendix
\newpage
\section{Proof of Theorem~\ref{theorem:coverage}}
\label{app:proof}

Note that Eq.~\ref{eq:estimates-rank-sets-coverage} holds if and only if:
\begin{equation} \label{eq:estimates-rank-sets-coverage-alternative}
    \lim_{n} \PP\left(\exists m \in \Mcal : \: [l(m),u(m)]\not\subseteq [\hat{l}(m),\hat{u}(m)]\right)\leq\alpha
\end{equation}
Therefore, to prove the theorem, it is sufficient to prove that Eq.~\ref{eq:estimates-rank-sets-coverage-alternative} 
holds.
Now, to prove that Eq.~\ref{eq:estimates-rank-sets-coverage-alternative}, we first show that the probability on the left hand side of the above equation is smaller than or equal to the probability $\PP(\thetab \notin \Ccal_{\alpha})$.

To this end, first note that, if for at least one model $m \in \Mcal$, we have that $\hat{l}(m)>l(m)$ or 
$\hat{u}(m)<u(m)$, then it holds that
\begin{equation*}
\bigcap_{m \in \Mcal}[l(m), u(m)] \not\subseteq [\hat{l}(m), \hat{u}(m)]. 
\end{equation*}
Next, without loss of generality, assume that, for model $m$, we have that $\hat{l}(m)>l(m)$.
In this case, from Eqs.~\ref{eq:estimates-rank-sets} and~\ref{eq:lower-upper-rank} we get:
\begin{equation*}
    \sum_{\tilde{m} \in \Mcal\setminus\{m\}}\one\{ d(\Ccal_{\alpha}, H_{m,\tilde{m}}) > 0\} \cdot \one\{ \thetahat_{m} < \thetahat_{\tilde{m}}\} \quad >
    \sum_{\tilde{m} \in \Mcal\setminus \{m\}}\one\{\theta_{m} < \theta_{\tilde{m}} \},
\end{equation*}
which means that there must be at least one model $\tilde{m} \in \Mcal$ such that %
$x_m<x_{\tilde{m}} \; \forall \xb\in \Ccal_{\alpha}$ and $\theta_m>\theta_{\tilde{m}}$, 
which implies that $\thetab \notin \Ccal_{\alpha}$.
As a result, we can immediately conclude that,
\begin{equation*}
\lim_{n} \PP\left(\exists m \in \Mcal : \: [l(m),u(m)]\not\subseteq [\hat{l}(m),\hat{u}(m)]\right) \, \leq \, \lim_{n} \PP(\thetab \notin \Ccal_{\alpha}) \, = \, \alpha.
\end{equation*}
This concludes the proof.\hfill\QED

\newpage
\section{Algorithms}
\label{app:algorithms}
\subsection{Algorithm~\ref{alg:1b}}
\label{app:alg-conf}

In this section, we present a pseudocode implementation of the algorithm to construct
confidence regions using linear regression.

\begin{algorithm}[h]
\DontPrintSemicolon
\small
\caption{It estimates $\hat{\thetab}$ and $\widehat{\Sigmab}$ using linear regression.}
\label{alg:1b}
\KwIn{$k$, $\Dcal$}
\KwOut{$\hat{\thetab},\widehat{\Sigmab}$}

\vspace{2mm}

$\bar{\wb}, \bar{\wb}',\Mb,\Mb' \gets$ \textsc{summarize}$(\Dcal,k)$

$\hat{\thetab} \gets \left(  \one_{k} \left(\left(\Mb+\Mb'\right)\one_{|\Dcal|}\right)^{\top}\odot \II_k \right)^{-1} \left( \Mb \cdot \bar{\wb} + \Mb' \cdot \bar{\wb}' \right)$ \tcp*{linear reg. estimate}

$\Ab \gets \left( \left( \one_{k}(\bar{\wb}-\Mb^{\top}\hat{\thetab})^{\top} \right)\odot \Mb + \left( \one_{k}(\bar{\wb}'-\Mb'^{\top}\hat{\thetab})^{\top} \right)\odot \Mb' \right)$

$\widehat{\Sigmab} \gets \frac{1}{|\Dcal|^2} \Ab \Ab^{\top}$\,\tcp*{estimate of covariance}

\Return{$\hat{\thetab},\widehat{\Sigmab}$} 
\end{algorithm}

Note that, in Algorithm~\ref{alg:2}, $\hat{\thetab}$ and $\widehat{\Sigmab}$ can alternatively be computed by calling Algorithm~\ref{alg:1b} instead of Algorithm~\ref{alg:1}.
Algorithm~\ref{alg:1b} runs in $O(k^2|\Dcal|)$ time.

\subsection{Algorithm to set $\lambda$}
\label{app:lambda}
The parameter $\lambda \in [0,1]$ weighs the comparisons $\hat{\wb},\hat{\wb}'$ by the strong LLM differently than the comparisons $\wb, \wb'$ by humans.
This way, if the strong LLM's preferences are strongly aligned with human preferences, the pairwise comparisons by the strong LLM can be weighed close to or equally with the pairwise comparisons by humans.
Conversely, if the strong LLM's preferences are not well aligned with human preferences, the pairwise comparisons $\hat{\wb},\hat{\wb}'$ can be weighed lower than the pairwise comparisons $\wb, \wb'$.
Following Angelopoulous et al.~\cite{angelopoulos2023ppi++}, we select the $\lambda$ that minimizes $\text{tr}(\Sigmab)$, where $\Sigmab=\EE[ (\hat{\thetab} - \thetab) (\hat{\thetab} - \thetab)^{\top} ]$:
\begin{equation}
    \lambda=\frac{n}{n+N}\frac{\text{tr}(\Sigma_n)} {\text{tr}(\Sigma_N)}
\end{equation}
where $\Sigma_N$ is the sample covariance matrix of the estimate of $\breve{\thetab}$ computed from Algorithm~\ref{alg:1b} using the pairwise comparisons by the strong LLM $\Dcal_N$,
and $\Sigma_n$ is the sample covariance matrix of the estimates of $\breve{\thetab}$ and $\thetab$ computed via Algorithm~\ref{alg:1b} using the pairwise comparisons by the strong LLM and by humans respectively in dataset $\Dcal_n$. Detailed computation of $\lambda$ is shown in Algorithm~\ref{alg:lambda}, which runs in $O(k^2(N+n))$.

\begin{algorithm}[h]
\DontPrintSemicolon
\small
\caption{It computes $\lambda$}
\label{alg:lambda}

\KwIn{$k$, $\Dcal_N$, $\Dcal_n$}
\KwOut{$\lambda$}

$\wb_n,\wb'_n,\hat{\wb}_n,\hat{\wb}'_n,\Mb_n,\Mb'_n \gets$\textsc{summarize}$(\Dcal_n,k)$


$\hat{\ab}_n \gets \left(  \one_{k} \left(\left(\Mb_{n}+\Mb'_{n}\right)\one_{n}\right)^{\top}\odot \II_k \right)^{-1} \left( \Mb_n \cdot \hat{\wb}_n + \Mb'_n \cdot\hat{\wb}'_n \right)$

$\ab_n \gets \left(  \one_{k} \left(\left(\Mb_{n}+\Mb'_{n}\right)\one_{n}\right)^{\top}\odot \II_k \right)^{-1} \left( \Mb_n \cdot \wb_n + \Mb'_n \cdot\wb'_n \right)$


$\hat{\Ab}_n \gets \left( \left( \one_{k}(\hat{\wb}_{n}-\Mb^{\top}_{n}\hat{\ab}_n)^{\top} \right)\odot \Mb_{n} + \left( \one_{k}(\hat{\wb}'_{n}-\Mb'^{\top}_{n}\hat{\ab}_n)^{\top} \right)\odot \Mb'_{n} \right)$

$\Ab_n \gets \left( \left( \one_{k}(\wb_{n}-\Mb^{\top}_{n}\ab_n)^{\top} \right)\odot \Mb_{n} + \left( \one_{k}(\wb'_{n}-\Mb'^{\top}_{n}\ab_n)^{\top} \right)\odot \Mb'_{n} \right)$


$\Sigma_n  \gets   \frac{1}{n^2}\hat{\Ab}_n \Ab_n^{-1}$ 

$-, \Sigma_N \gets \textsc{confidence-ellipsoid}(k,\Dcal_N)$ \tcp*{Algorithm~\ref{alg:1b}}

$\lambda \gets \frac{n}{n+N}\frac{\text{tr}(\Sigma_n)} {\text{tr}(\Sigma_N)}$

\Return{$\lambda$}
\end{algorithm}

\newpage
\section{Additional Details of the Experimental Setup}
\label{app:experimental-setup}
In this section, we provide the implementation details and computational resources used to execute the experiments discussed in Section~\ref{sec:experiments}. Our algorithms are implemented in {\tt Python 3.11.2} programming language using {\tt NumPy} and {\tt SciPy} open-source libraries for efficient matrix operations. Further, we use the {\tt matplotlib} package to facilitate visualizations of our results. The complete details of the software requirements can be found in the source code provided as part of the supplementary materials. Our experiments are executed on a compute server equipped with $2\,\times$ AMD EPYC 7702 processor with $64$ cores per processor and $2$ TB of main memory. It is worth noting that, our experiments are not resource intensive and can be executed on a standard desktop computer or a laptop.

\begin{figure}
\centering
\includegraphics[width=0.7\textwidth]{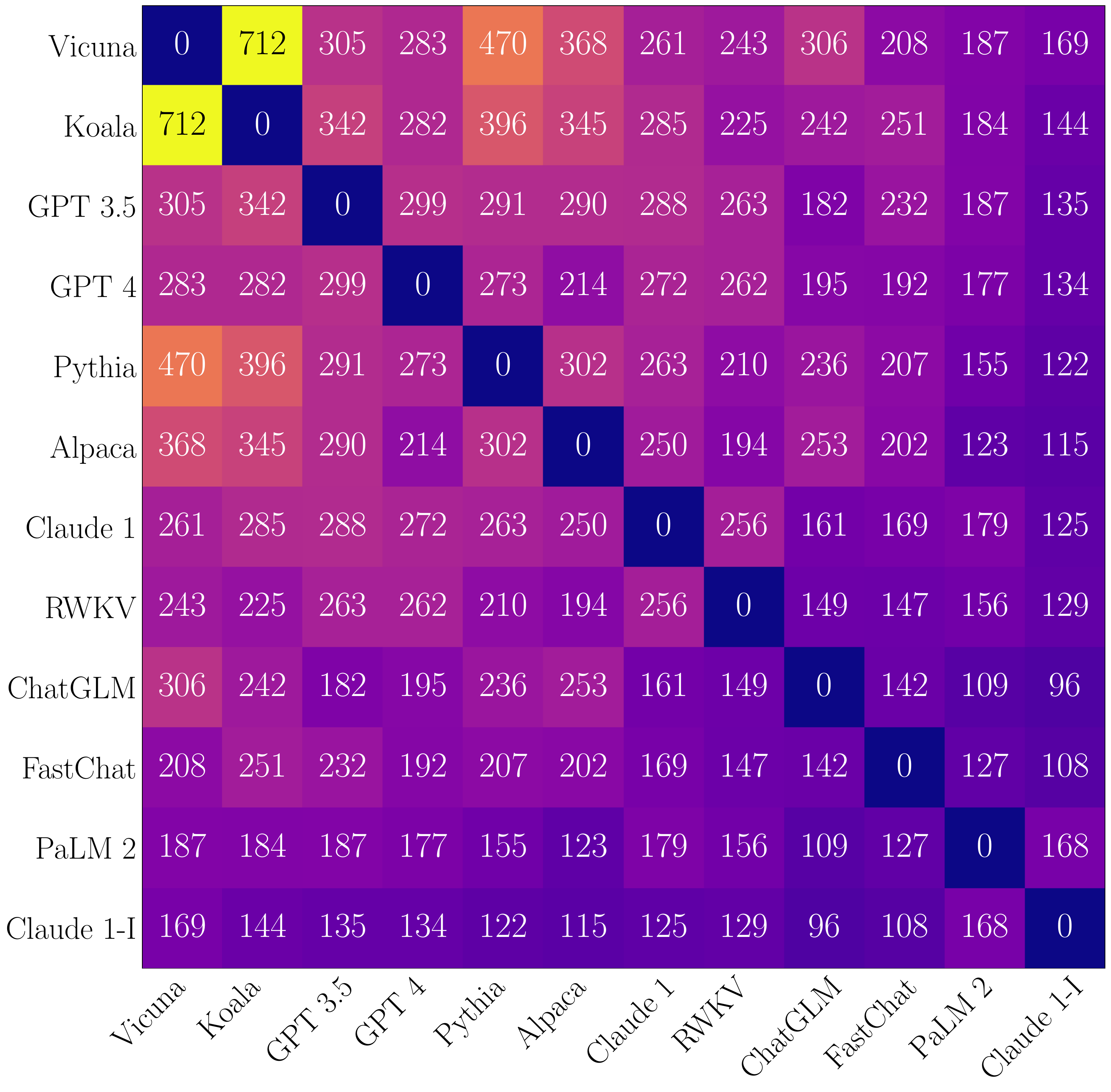}
\caption{The number of pairwise comparisons per each pair of models after all preprocessing steps.}
\label{fig:pairwise-comparisons}
\end{figure}

\xhdr{Pairwise comparisons and preprocessing}
Using the dataset from LMSYS Chatbot Arena\footnote{The user prompts are licensed under CC-BY-4.0, while the model outputs are licensed under CC-BY-NC-4.0.}, we gathered pairwise comparisons of three strong LLMs via API calls to {\tt OpenAI API version 2024\-02-15-preview} for GPT 3.5 and GPT 4, and {\tt Anthropic API version 2023-06-01} for Claude 3. To this end, we use (almost) the same prompt as in Zheng et al.~\cite{zheng2023judging} that instructs each strong LLM to output option `A' (`B') if it prefers the response of first (second) model, or option `C' if it declares a tie. The prompt used to obtain pairwise preferences of is available in Listing~\ref{list:pair-pref}.
We preprocess the dataset by filtering out instances---an instance is a pair of responses to a query by two different models and the query itself---for which at least one strong LLM returned a verbose output instead of `A', `B' or `C', and choose a set of $12$ popular large language models for our experiments. The chosen models, along with their specific versions, are listed in Table~\ref{tab:model_names}. In Figure~\ref{fig:pairwise-comparisons}, we show the number of pairwise comparisons per each pair of these chosen models after completing all preprocessing steps.

\begin{figure}[]
\begin{lstlisting}[label=list:pair-pref, caption= The prompt used for gathering pairwise preferences of strong LLMs.]
[System]
Act as an impartial judge and evaluate the responses of two AI
assistants to the user's question displayed below. Your evaluation
should consider factors such as relevance, helpfulness, accuracy,
creativity and level of detail of their responses. Output your final
verdict by strictly following this format: 'A' if the response of
assistant A is better, 'B' if the response of assistant B is better, 
and 'C' for a tie. Do not give any justification or explanation for 
your response.

[User]
Question:
{Query}

Response of assistant A:
{Response A}

Response of assistant B:
{Response B}
\end{lstlisting}
\end{figure}

\begin{table}[]
\caption{The names and versions of the $12$ popular large language models considered for our experiments after preprocessing the Chatbot Arena dataset.}
\label{tab:model_names}
\centering
{
\vspace{0.2cm}
\begin{tabular}{l l}
\hline
\textbf{Large Language Model} & \textbf{Version} \\ [0.5ex] 
\hline
    GPT 4      & GPT 4  \\  
    Claude 1   & Claude v1 \\
    Claude 1-I & Claude Instant v1  \\
    GPT 3.5    & GPT 3.5 turbo \\
    Vicuna     & Vicuna 13B \\
    PaLM 2     & PaLM 2 \\
    Koala      & Koala 13B \\
    RWKV       & RWKV 4 Raven 14B \\
    Pythia     & OpenAssistant Pythia 12B \\
    Alpaca     & Alpaca 13B \\
    ChatGLM    & ChatGLM 6B \\
    FastChat   & FastChat T5 3B \\
\hline
\end{tabular}
}
\end{table}

\newpage
\section{Additional Experimental Results}
\label{app:experimental-results}
In this section, we discuss additional experimental results that were omitted from the main paper due to space limitations.

\subsection{Quality of the Rank-sets}\label{app:confidence-intervals-intersection}
\begin{figure}[t]
\subfloat[GPT 4]{\includegraphics[width=1.0\textwidth]{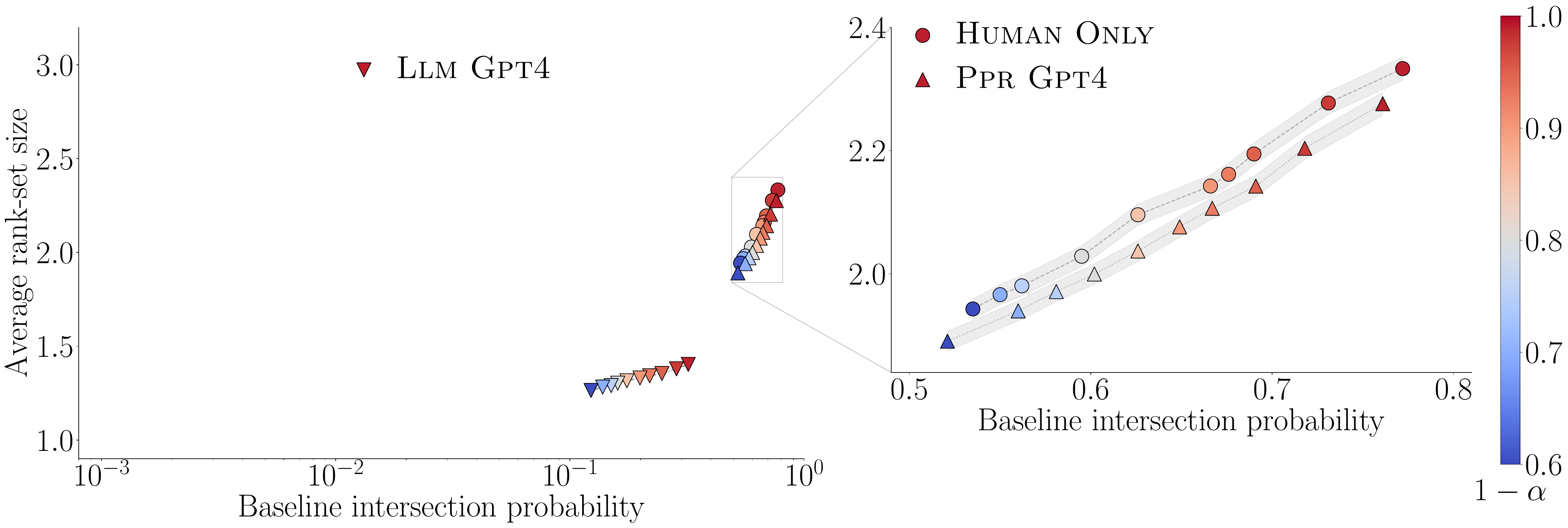}}\\
\subfloat[Claude 3]{\includegraphics[width=1.0\textwidth]{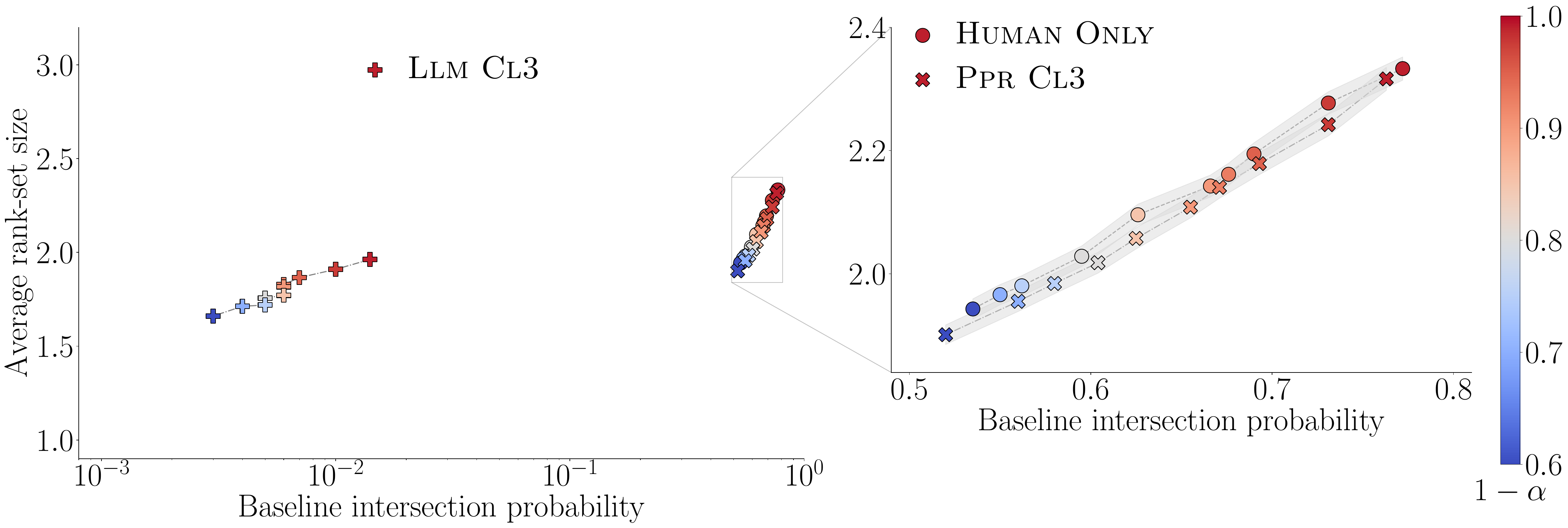}}\\
\subfloat[GPT 3.5]{\includegraphics[width=1.0\textwidth]{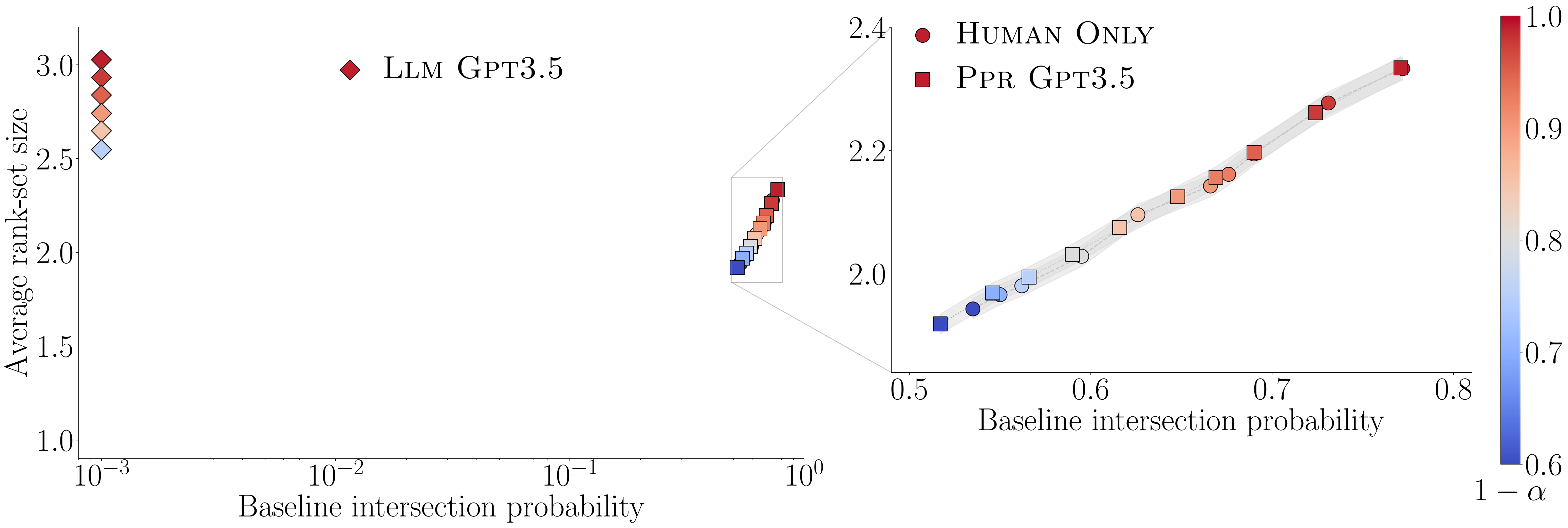}} \\
\caption{
Average rank-set size against baseline intersection probability for rank-sets constructed using only pairwise comparisons by a strong LLM: \llm~\gptfour (top), \llm~\claude (middle) and \llm~\gptthree (bottom), 
only pairwise comparisons by humans (\human), 
and pairwise comparisons by both a strong LLM and humans (\ppr~\gptfour top, \ppr~\claude middle and \ppr~\gptthree bottom) for different $\alpha$ values and $n=990$. 
Smaller (larger) average rank-set sizes and larger (smaller) intersection probabilities are better (worse).
The shaded region shows a $95\%$ confidence interval for the rank-set size of all rank-sets among all $1,000$ repetitions.
\label{fig:rankset-size-vs-intersection-probability-2}}
\end{figure}

In Figure~\ref{fig:rankset-size-vs-intersection-probability-2}, we show a more detailed analysis of the average rank-set size plotted against the baseline intersection probability, with individual plots for each strong LLM: \gptfour (top), \claude (middle) and \gptthree (bottom), for $n=990$ and different $\alpha$ values.\footnote{\label{foot:alpha}$\alpha \in \{0.4, 0.3, 0.25, 0.2, 0.15, 0.1, 0.075, 0.05, 0.025, 0.01\}$} All observations made in Figure~\ref{fig:rankset-size-vs-intersection-rate} apply to Figure~\ref{fig:rankset-size-vs-intersection-probability-2} as well. Moreover, it is evident that among the three strong LLMs, GPT 4 clearly outperforms the others, achieving higher baseline intersection probability and returning rank-sets with smaller average size. Specifically, in Figure~\ref{fig:rankset-size-vs-intersection-probability-2} (a), the difference between average rank-set size between \human and \ppr~\gptfour is significantly pronounced, while the gap gradually narrows in subsequent plots for \ppr~\claude and \ppr~\gptthree.

\subsection{Baseline Coverage Probability}\label{app:baseline-coverage-probability}
In this subsection, we investigate a more conservative quality metric, namely baseline coverage probability, which is the (empirical) probability that the rank-sets $[\hat{l}(m),\hat{u}(m)]$ constructed by any method cover the rank-sets $[\tilde{l}(m),\tilde{u}(m)]$ constructed using the \baseline method, \ie,
\begin{equation}
\PP \left( \bigcap_{m\in\Mcal}\one \left\{[\tilde{l}(m),\tilde{u}(m)] \subseteq [\hat{l}(m),\hat{u}(m)]\right\} \right)
\end{equation}
The baseline intersection probability, which we discussed in Section~\ref{sec:experiments} and illustrated in Figure~\ref{fig:rankset-size-vs-intersection-rate}, is a less conservative metric compared to the baseline coverage probability. While the latter represents the probability that all rank-sets are covered by the \baseline rank-sets, the former only considers the probability that the rank-sets intersect.
Thus, achieving high baseline coverage probability is difficult, particularly when the \baseline rank-sets are larger.

\begin{figure}[t]
\includegraphics[width=1.0\textwidth]{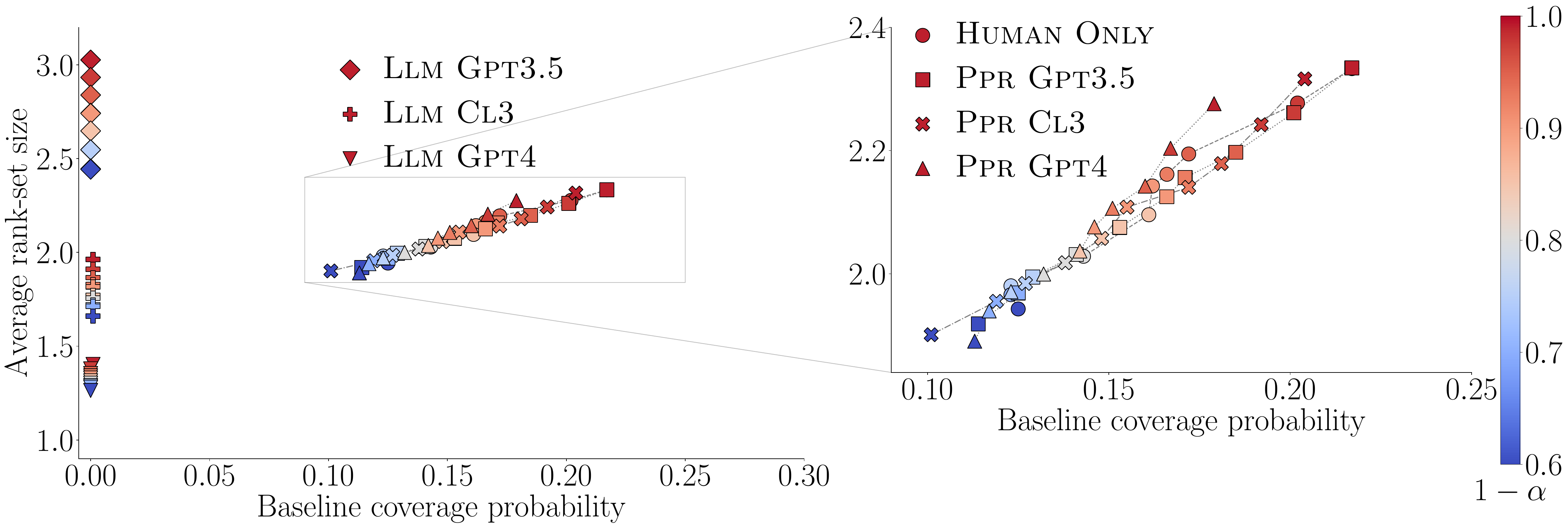}
\caption{Average rank-set size against baseline coverage probability for rank-sets constructed using only pairwise comparisons by a strong LLM (\llm~\gptfour, \llm~\gptthree and \llm~\claude), 
only pairwise comparisons by humans (\human), 
and pairwise comparisons by both a strong LLM and humans (\ppr~\gptfour, \ppr~\gptthree and \ppr~\claude) for different $\alpha$ values and $n=990$. 
Smaller (larger) average rank-set sizes and larger (smaller) coverage probabilities are better (worse). In all panels, $95\%$ confidence bars for the rank-set size are not shown, as they are below $0.02$.
\label{fig:coverage-size}}
\end{figure}

\xhdr{Quality of the rank-sets when considering the baseline coverage probability}
In Figure~\ref{fig:coverage-size}, we show the average rank-set size against the baseline coverage probability for rank-sets constructed using all methods (except \baseline) for $n=990$ and different values of $\alpha$ (the same values as in Figure~\ref{fig:rankset-size-vs-intersection-probability-2}).
Immediately, we notice that the baseline coverage probability of all methods is very low.
For rank-sets constructed using pairwise comparisons only by a strong LLM (\llm~\gptfour, \llm~\claude and \llm~\gptthree), the baseline coverage probability is close to (or exactly) zero.
Rank-sets constructed using only pairwise comparisons by humans (\human) or prediction-powered ranking using a strong LLM (\ppr~\gptfour, \ppr~\claude and \ppr~\gptthree) achieve better baseline coverage probability.
However, the difference in performance among these methods is not clear to distinguish, which motivated us to consider the baseline intersection probability metric for our experimental results in Section~\ref{sec:experiments}.

\begin{figure}[t]
\includegraphics[width=1\textwidth]{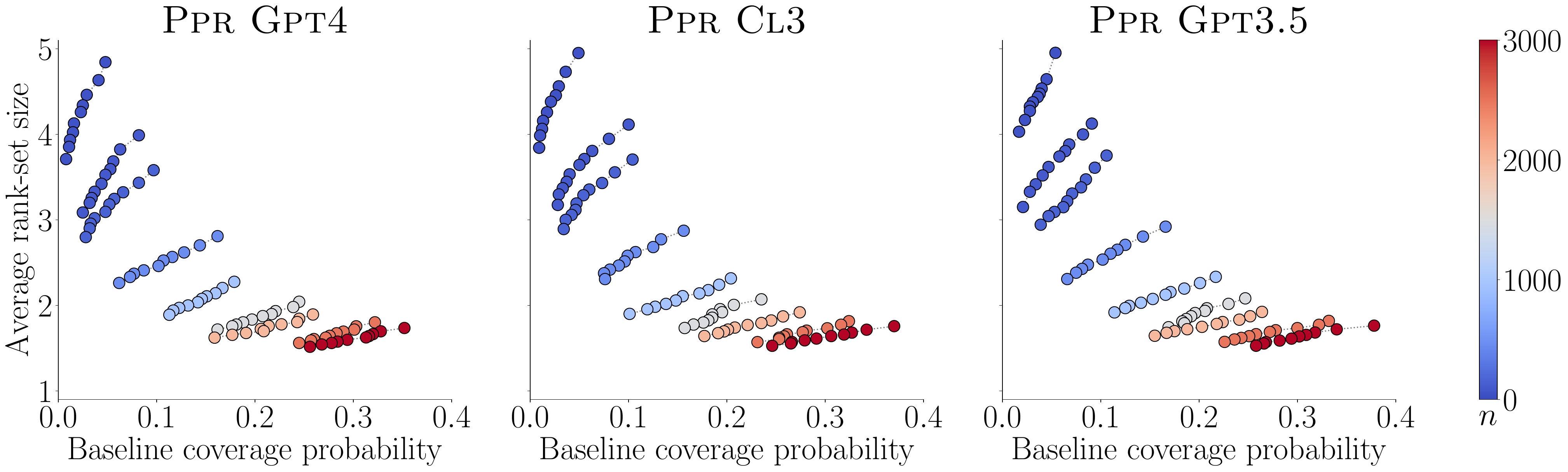}
\caption{Average rank-set size against baseline coverage probability for rank-sets constructed using pairwise comparisons by both a strong LLM and humans for different $n$ and $\alpha$ values. Smaller (larger) average rank-set sizes and larger (smaller) coverage probabilities are better (worse). In all panels, $95\%$ confidence bars for the rank-set size are not shown, as they are below $0.04$.
\label{fig:coverage-vary-n}}
\end{figure}

In Figure~\ref{fig:coverage-vary-n}, we show the average rank-set size against the baseline coverage probability for rank-sets constructed using \ppr~\gptfour, \ppr~\gptthree and \ppr~\claude 
for different values
\footnote{$n \in \{66, 132, 198, 462, 990, 1452, 1980, 2442, 2970\}$.} of $n$ and $\alpha$ (the same $\alpha$ values as in Figure~\ref{fig:rankset-size-vs-intersection-probability-2}).
Similarly as in Figure~\ref{fig:vary-n-ppr}, results show that there is a trade-off between average rank-set size and the baseline coverage probability, which improves rapidly as the number of pairwise comparisons by humans $n$ increases, but with diminishing returns.

\clearpage
\newpage

\subsection{Structure of the Rank-sets} \label{app:structure-of-ranksets}
\begin{figure}[t] 
\includegraphics[width=1\textwidth]{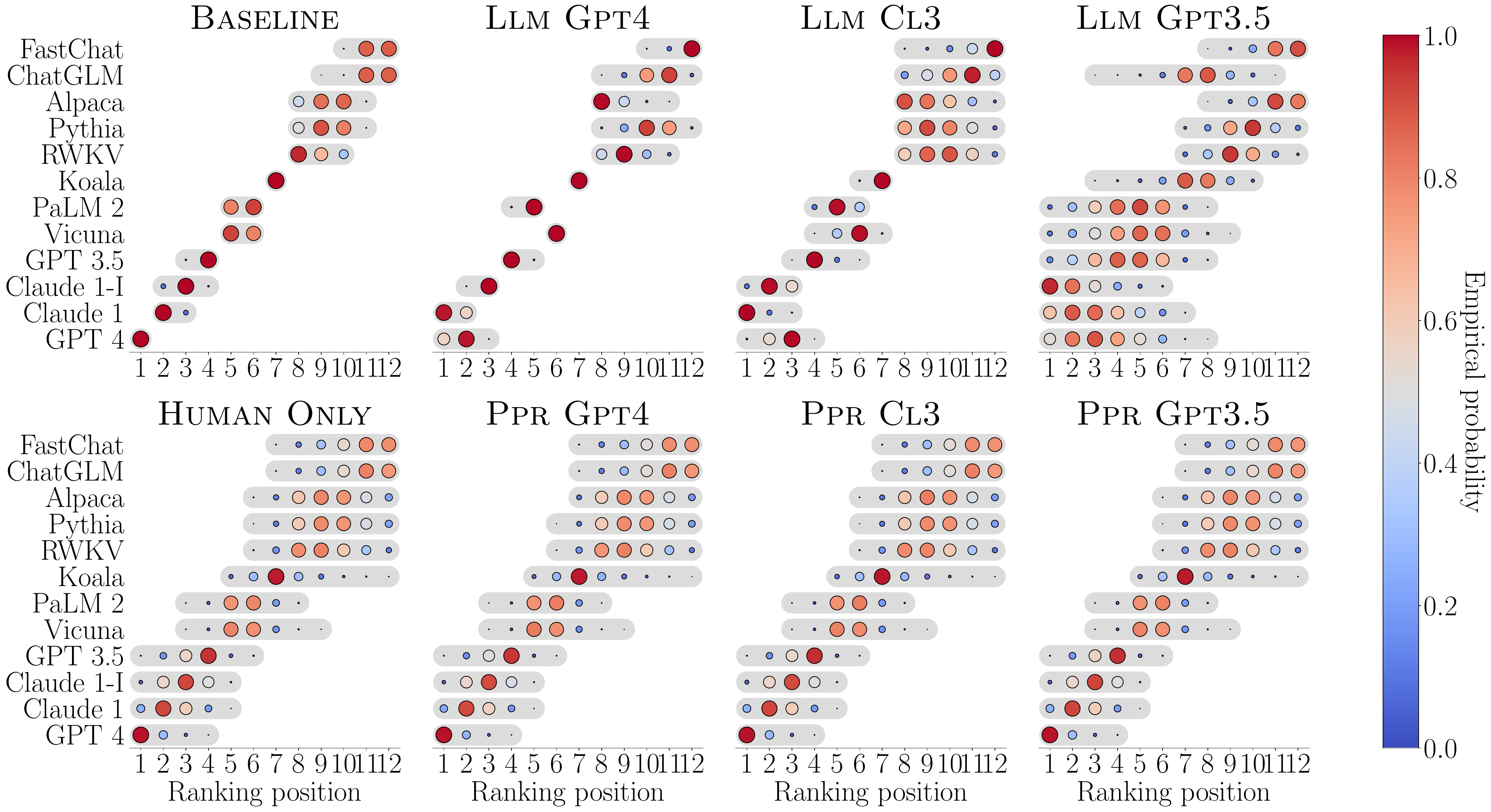}
\caption{\label{fig:ranksets-all-methods} Empirical probability that each ranking position is included in the rank-sets constructed by all methods for each of the LLMs under comparison. In all panels, $n=990$ and $\alpha=0.05$.
Larger (smaller) dots indicate higher (lower) empirical probability.} \label{fig:ranks-all}
\end{figure}

In this subsection, we compute the empirical probability that each ranking position is included in the rank-sets constructed by all methods, of each of the LLMs under comparison. The results are summarized in Figure~\ref{fig:ranks-all} for $\alpha=0.05$ and $n=990$.

\xhdr{Empirical probability of ranking positions} Our first observation is that, the ranking positions of each model in \llm~\gptfour and \llm~\claude have lower uncertainty compared to \ppr~\gptthree and \ppr~\claude, respectively. However, \llm~\gptthree exhibits higher uncertainty compared to \ppr~\gptthree. Nonetheless, the ranking positions with the highest probability mass in \llm~\gptfour, \llm~\claude and \llm~\gptthree significantly deviate from the \baseline. Specifically, the ranking position with highest probability mass differs for $7$ out of $12$ models.
In contrast, for \ppr~\gptfour, \ppr~\claude and \ppr~\gptthree it only differs from \baseline for $3$ out of $12$ LLMs.
These findings once again question the rationale of relying solely on pairwise comparisons by strong LLMs to rank LLMs~\cite{wang2023large,chiang2023vicuna,chiang2023can,jiang2023llm,wang2024pandalm,dubois2024alpacafarm,zheng2023judging}.
Our second observation is that there is no significant difference in the uncertainty in ranking positions across \human, \ppr~\gptfour, \ppr~\claude and \ppr~\gptfour. However, the distribution of probability mass across different ranking positions differs slightly among these methods, as clearly seen in \ppr~\gptfour, where the Alpaca model has zero probability mass for position $6$. 

\xhdr{Empirical probability of rank-sets}
Next, we compute the empirical probability of each rank-set constructed by all methods and for each of the LLMs under comparison with $n=990$ and $\alpha=0.05$. The results are summarized in Figure~\ref{fig:all-ranksets}.
Consistent with the observations from Figure~\ref{fig:ranksets-all-methods}, we note that the distribution of rank-sets constructed by \llm~\gptfour is more concentrated than those constructed by other methods. Conversely, \llm~\gptthree exhibits a more spread-out distribution of rank-sets, indicating higher uncertainty in its ranking positions. This observation is consistent across all LLMs considered for ranking.
Moreover, for \llm~\gptfour and \llm~\claude, we observe that the rank-sets with the highest probability mass often differed from those of the \baseline, with discrepancies observed in $7$ out of $12$ models.
On the contrary, for \ppr~\gptfour, \ppr~\claude and \ppr~\gptthree, we observe that the rank-sets with the highest probability mass match those constructed by \baseline more frequently than their \llm only counterparts.
Once again, these findings suggest that rank-sets obtained using only pairwise comparisons of strong LLMs are not reliable.

\begin{figure}
\subfloat{\includegraphics[width=1.0\textwidth]{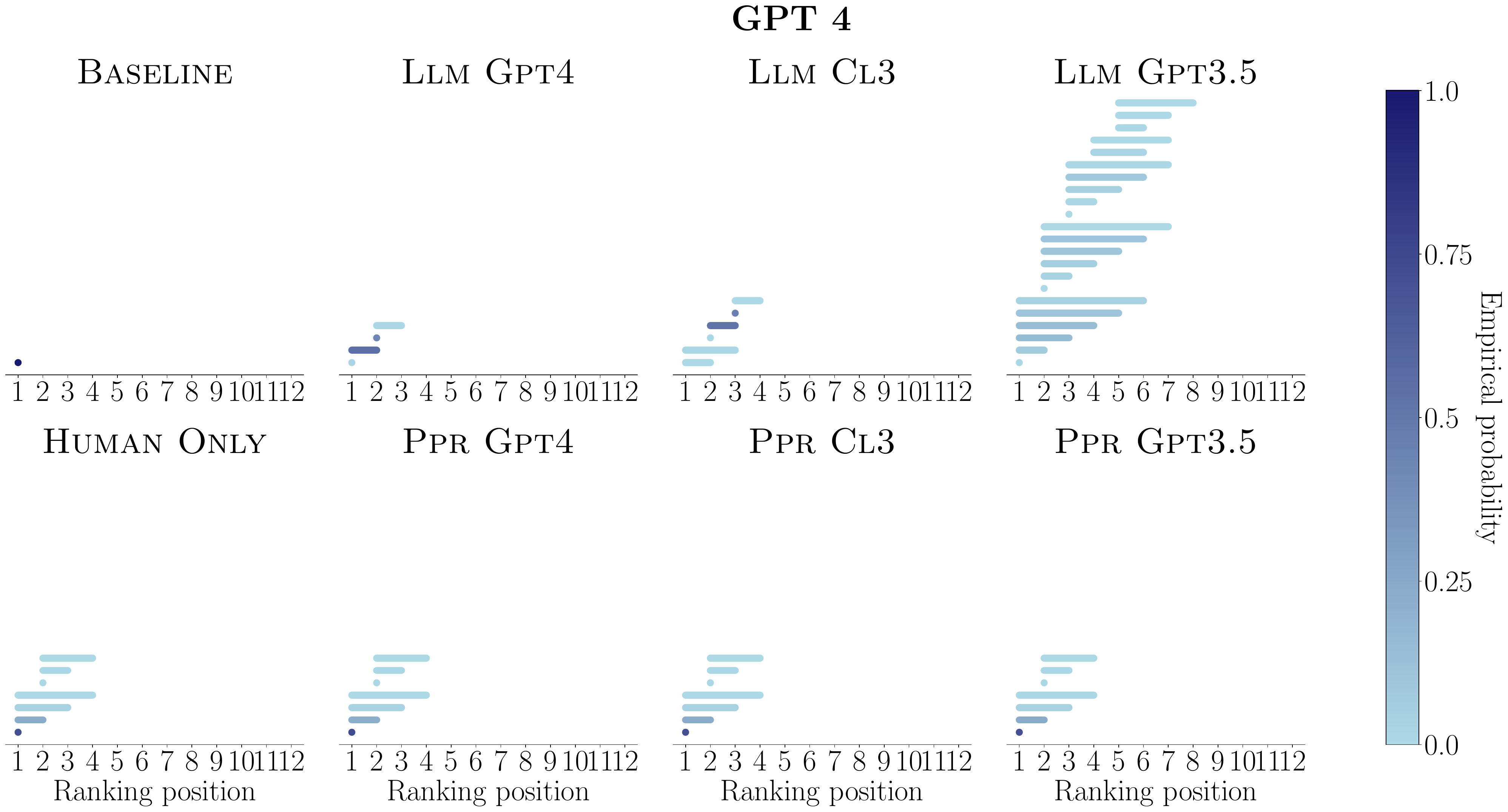}}\\
\subfloat[Claude 1]{\includegraphics[width=1.0\textwidth]{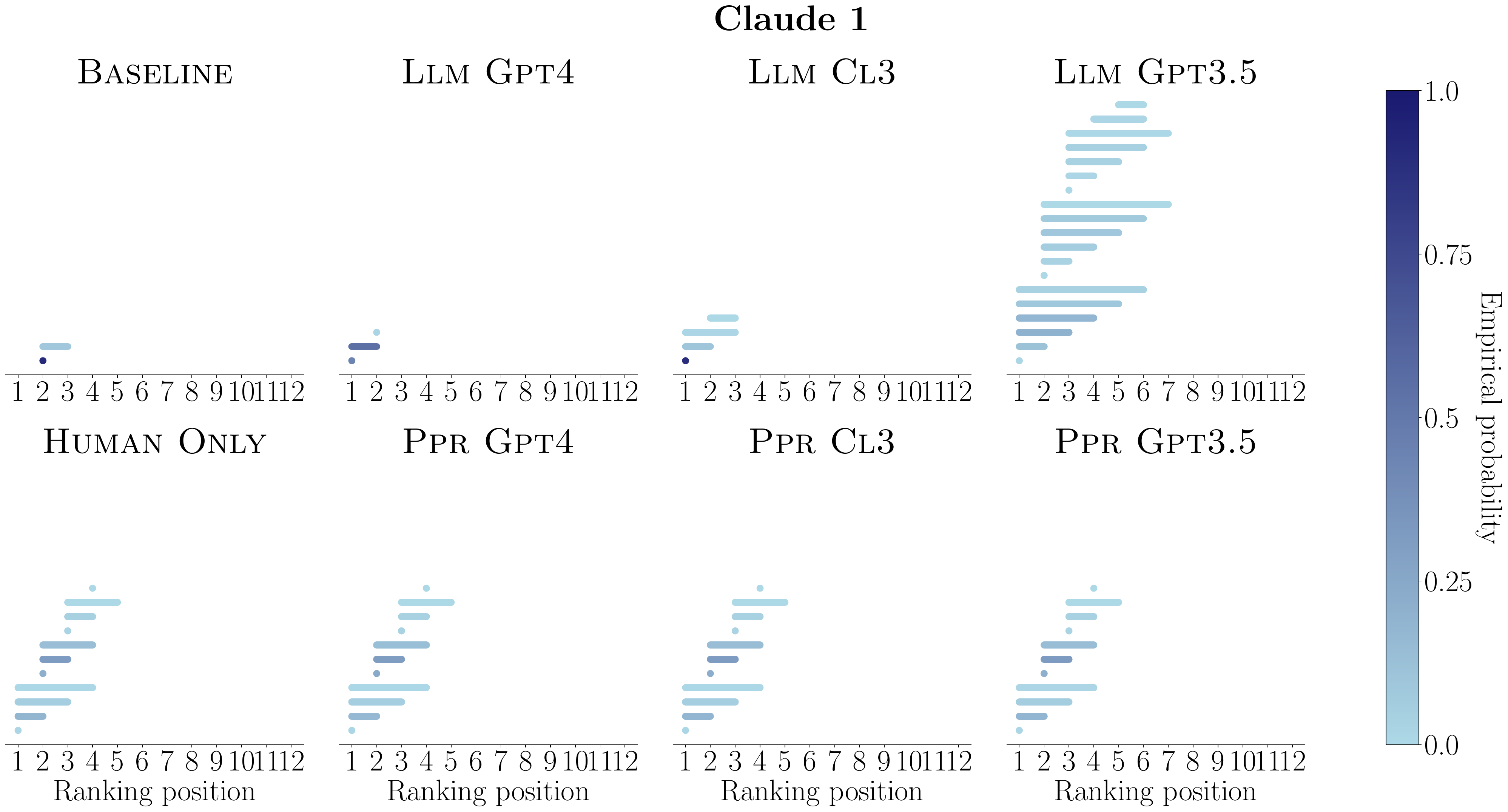}}
\caption{Empirical probability of each rank-set constructed by all methods for all 12 models (one model per sub-figure).
In all panels, $n=990$ and $\alpha=0.05$.}
\end{figure}
\begin{figure}
\ContinuedFloat
\subfloat[Claude 1-I]{\includegraphics[width=1.0\textwidth]{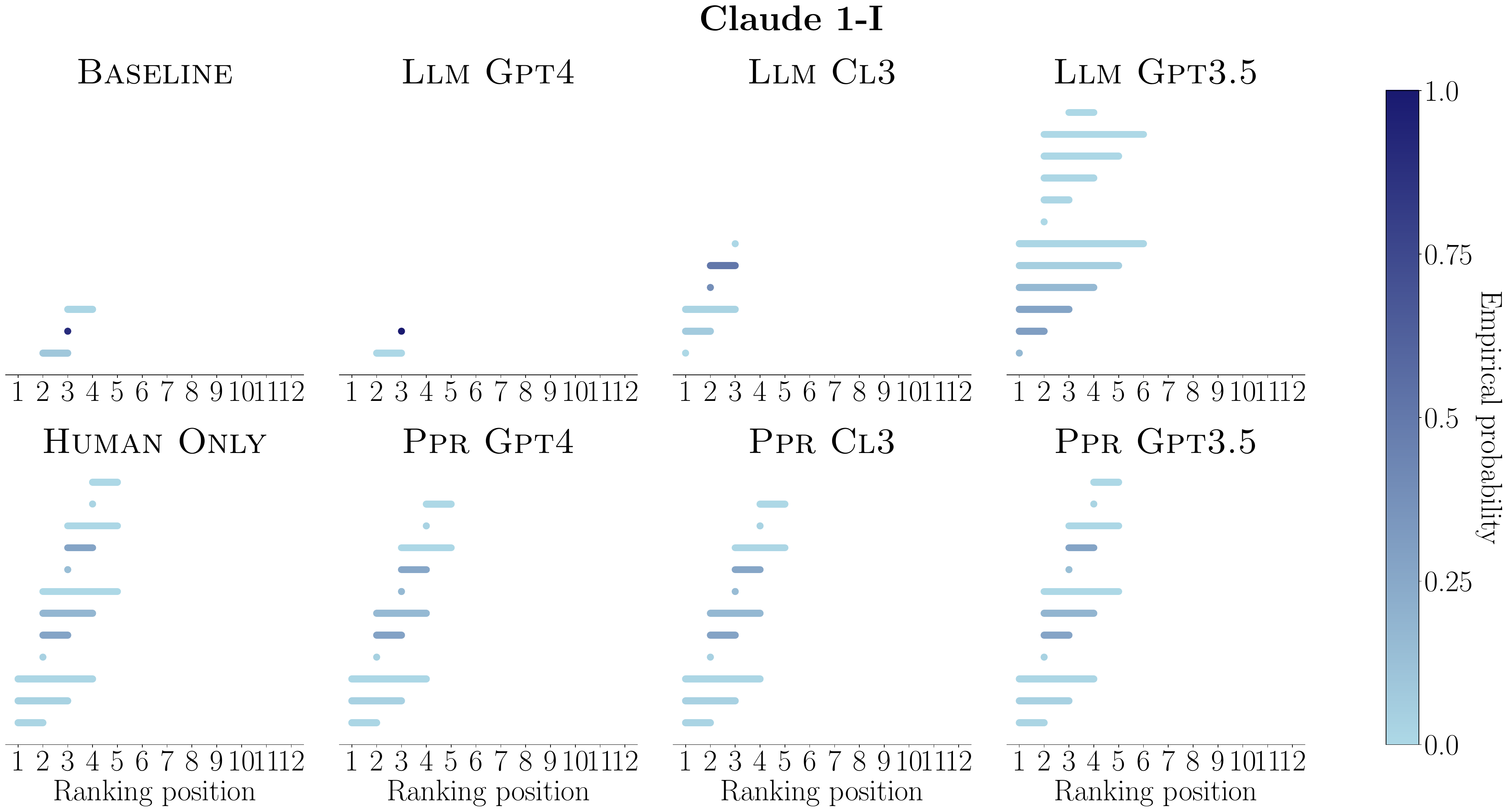}}\\
\subfloat{\includegraphics[width=1.0\textwidth]{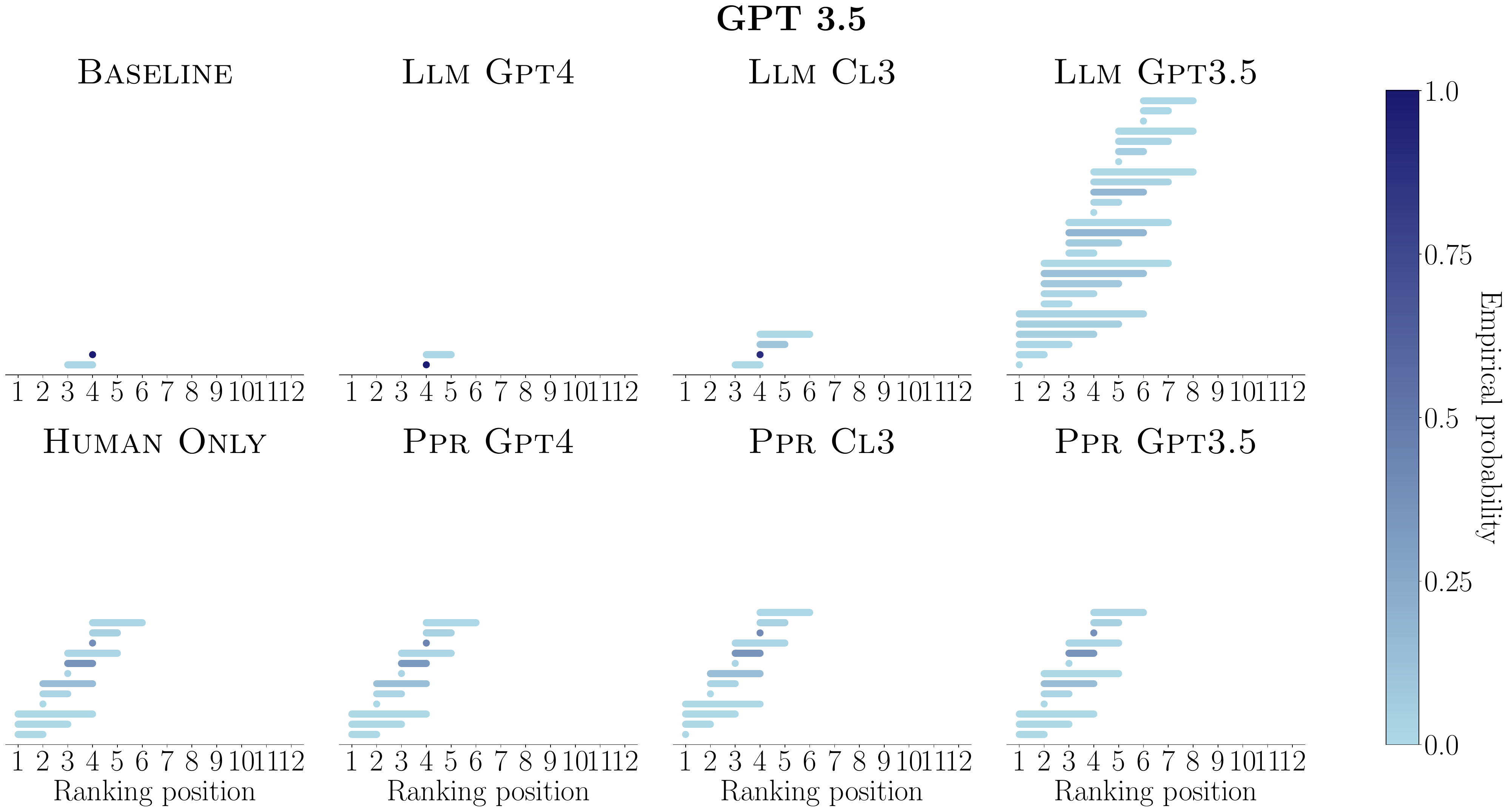}}
\captionsetup{list=off,format=cont}
\caption{}
\end{figure}
\begin{figure}\ContinuedFloat
\subfloat{\includegraphics[width=1.0\textwidth]{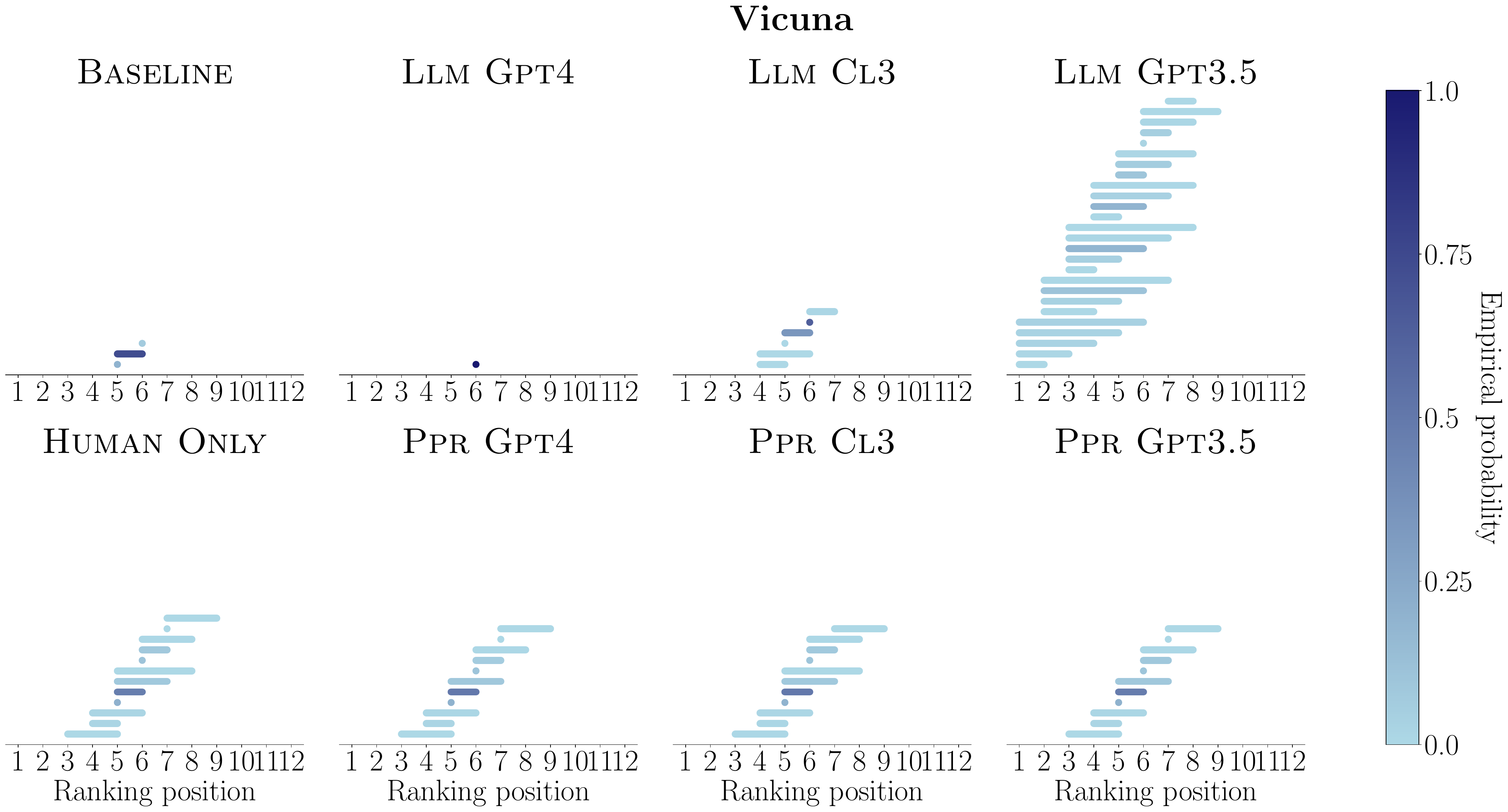}}\\
\subfloat{\includegraphics[width=1.0\textwidth]{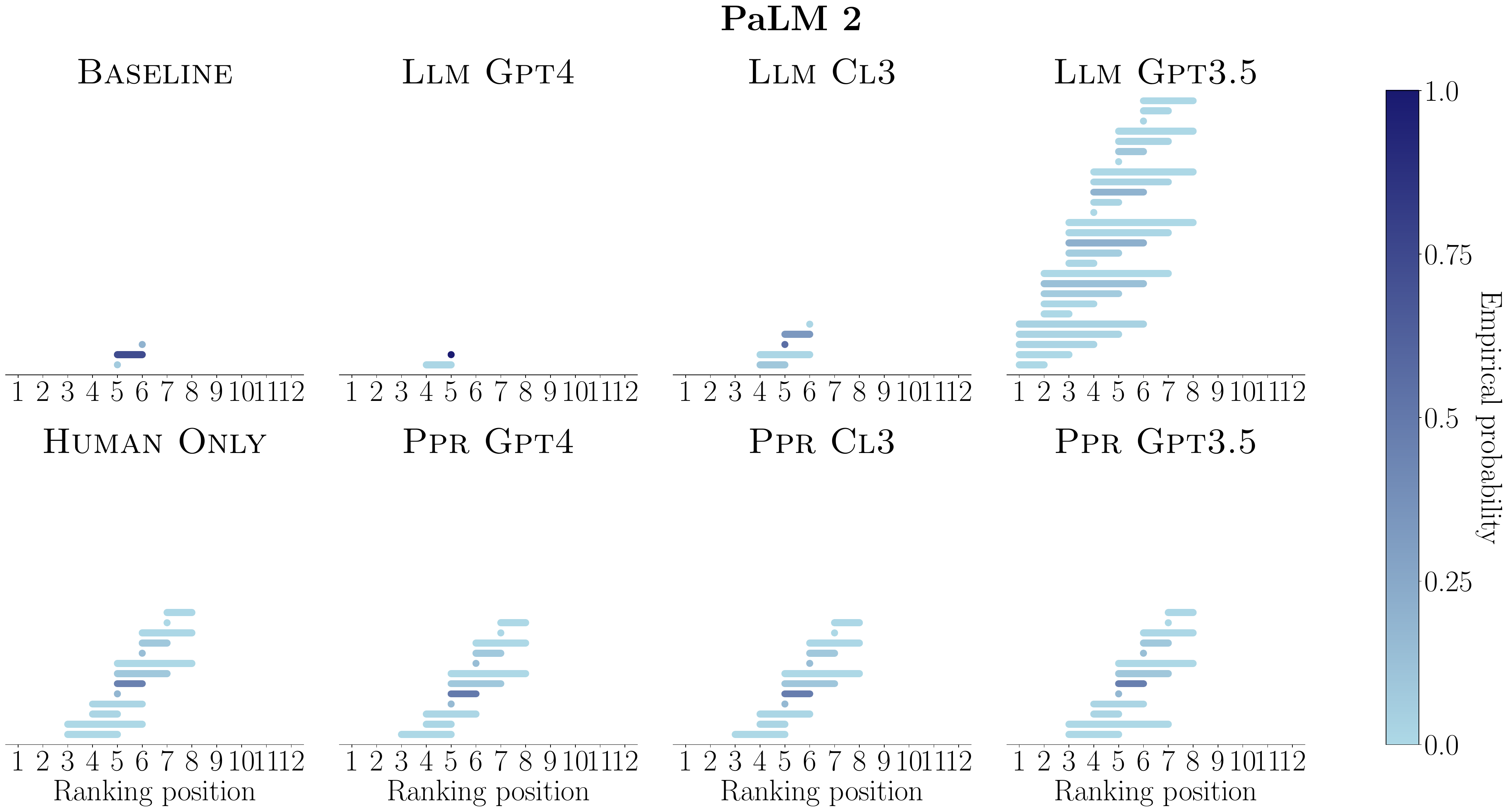}}
\captionsetup{list=off,format=cont}
\caption{}
\end{figure}
\begin{figure}\ContinuedFloat
\subfloat{\includegraphics[width=1.0\textwidth]{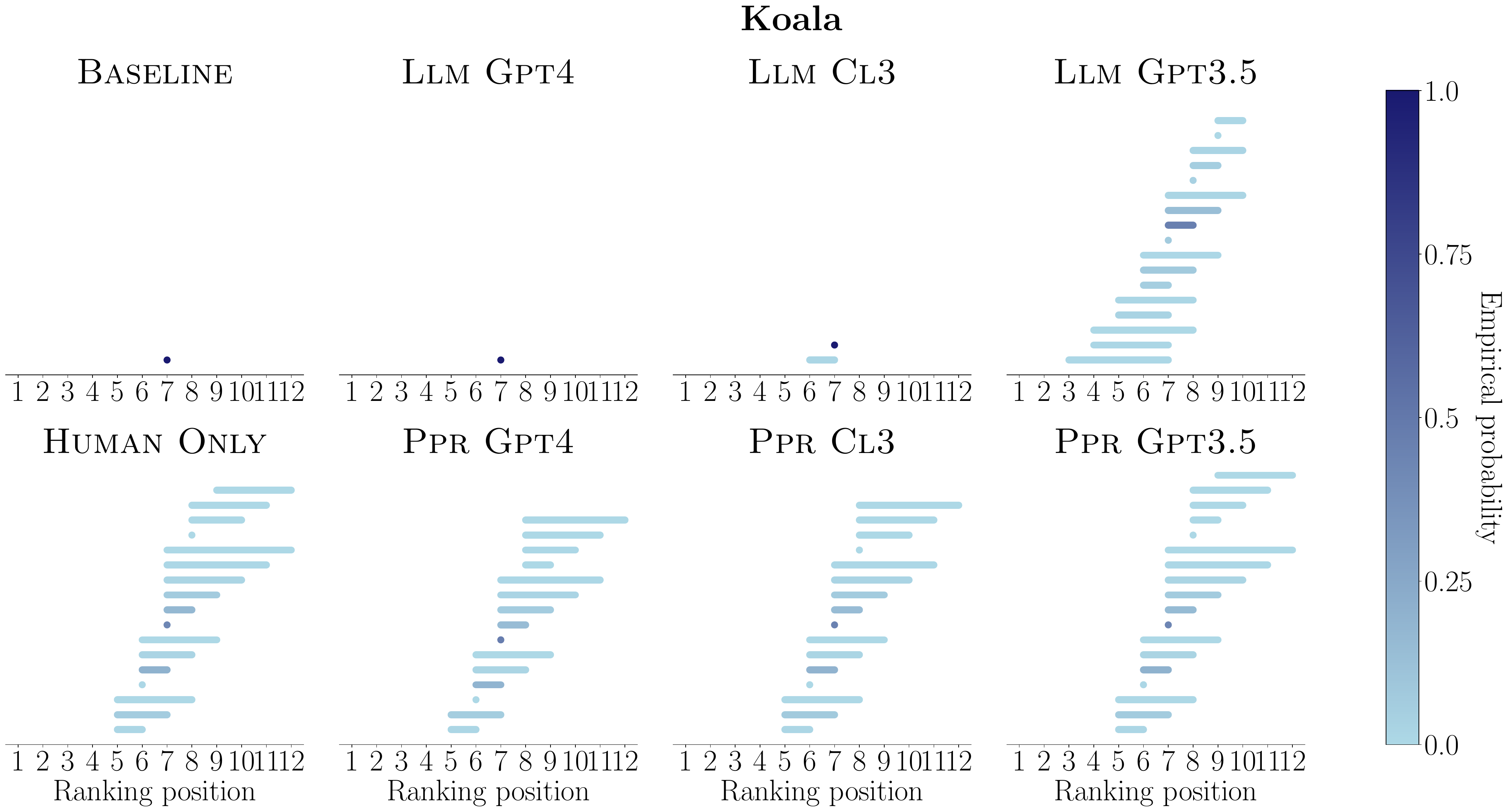}}\\
\subfloat{\includegraphics[width=1.0\textwidth]{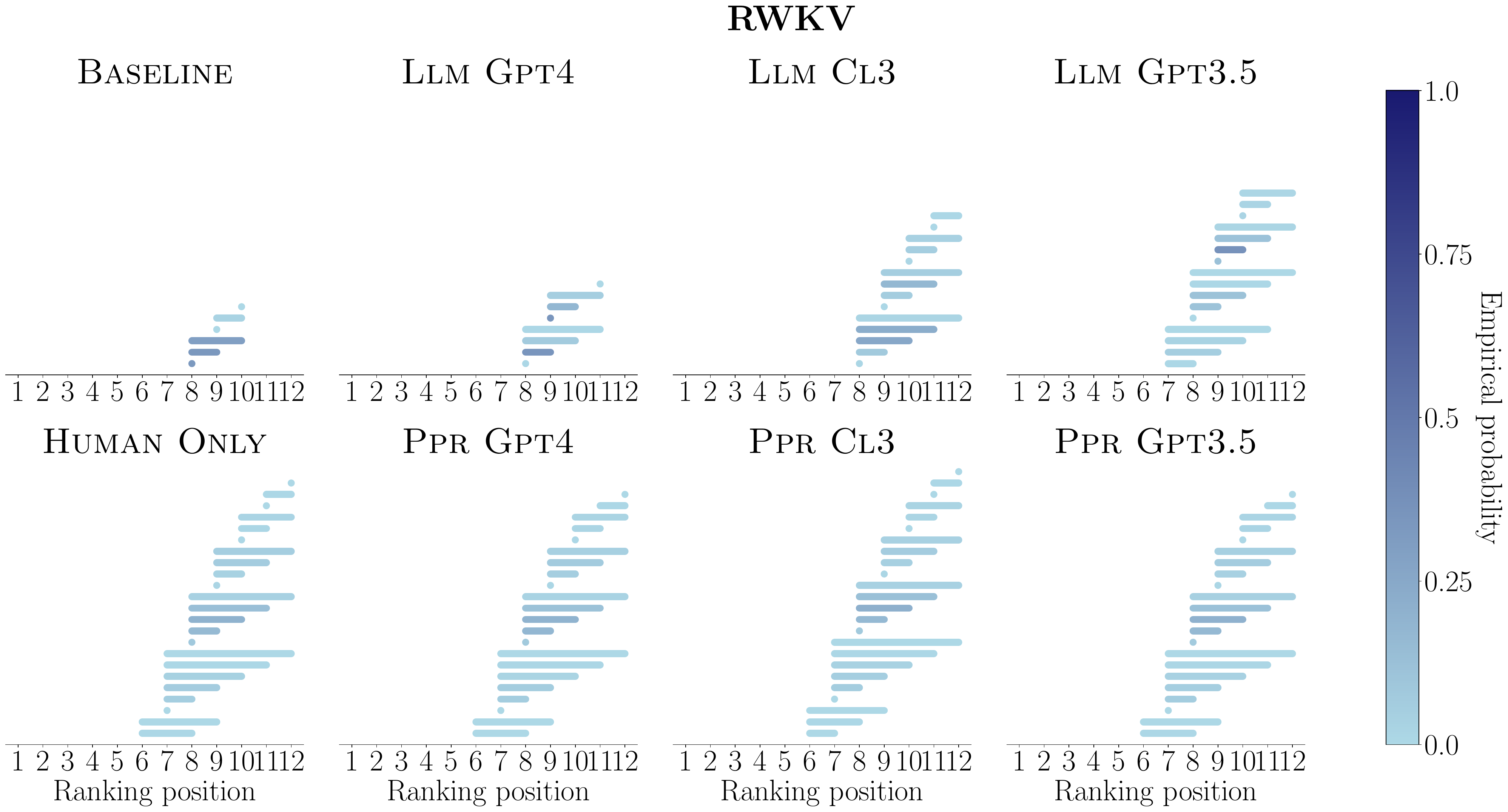}}
\captionsetup{list=off,format=cont}
\caption{}
\end{figure}
\begin{figure}\ContinuedFloat
\subfloat{\includegraphics[width=1.0\textwidth]{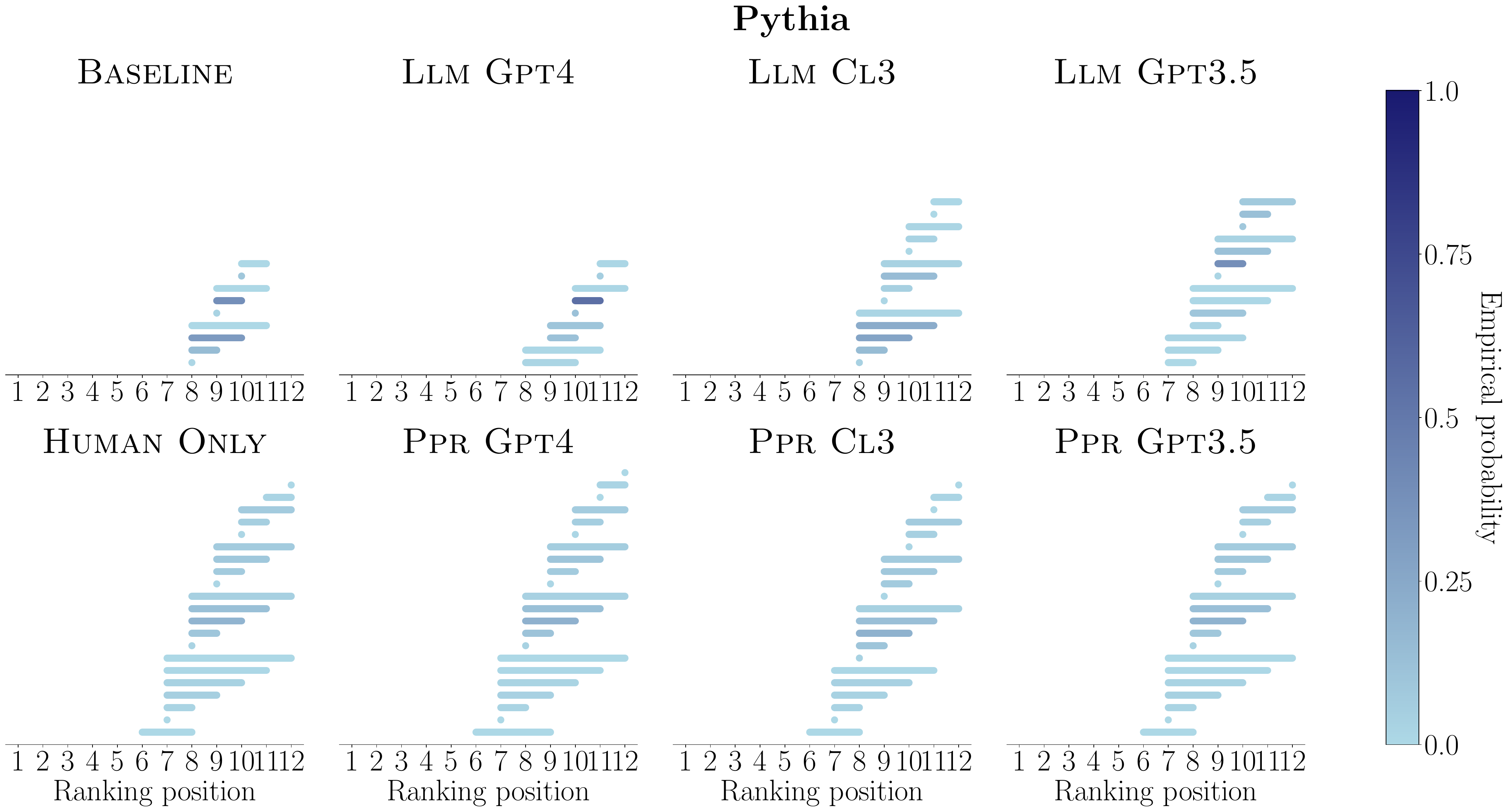}}\\
\subfloat{\includegraphics[width=1.0\textwidth]{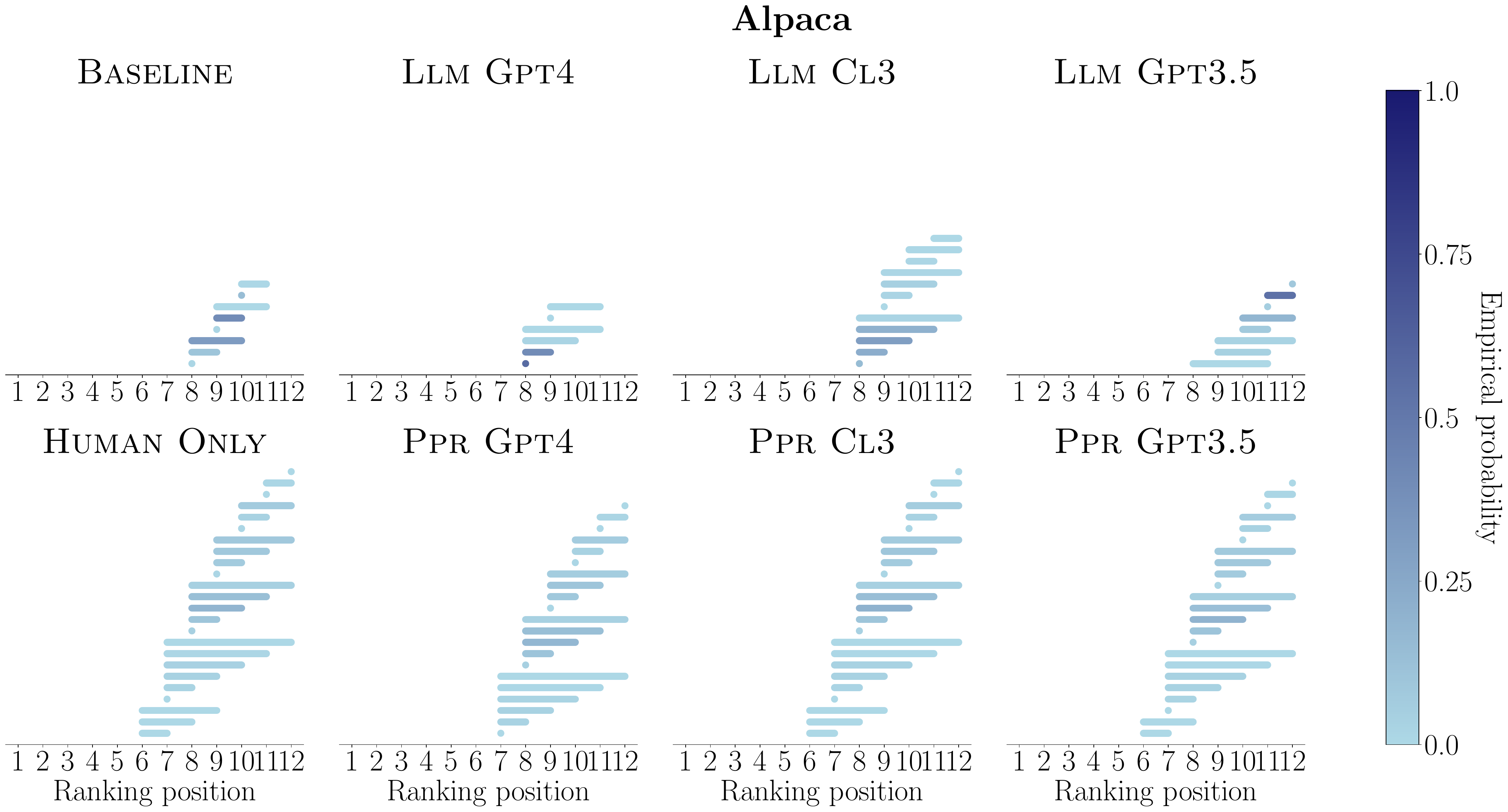}}
\captionsetup{list=off,format=cont}
\caption{}
\end{figure}
\begin{figure}\ContinuedFloat
\subfloat{\includegraphics[width=1.0\textwidth]{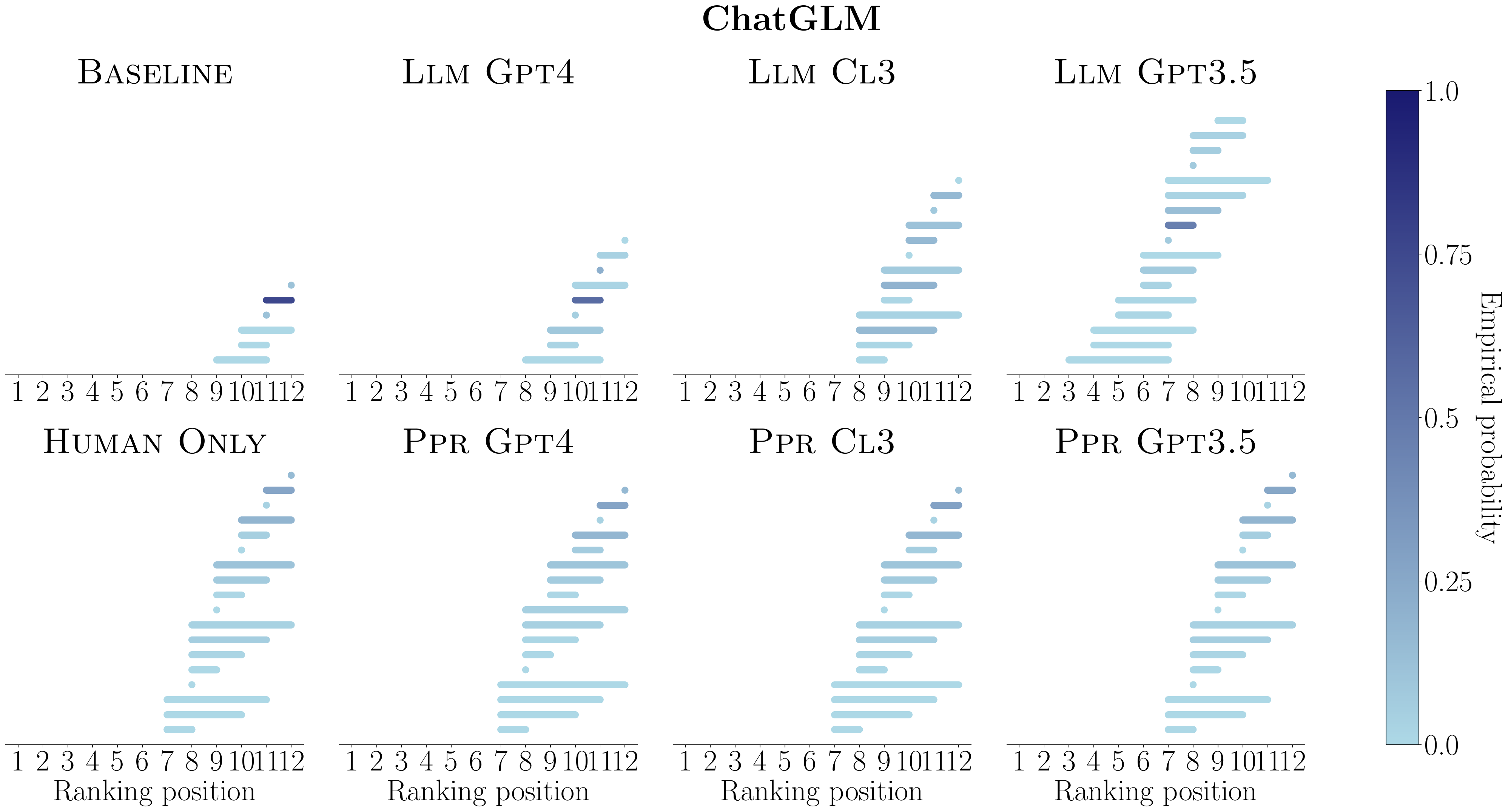}}\\
\subfloat{\includegraphics[width=1.0\textwidth]{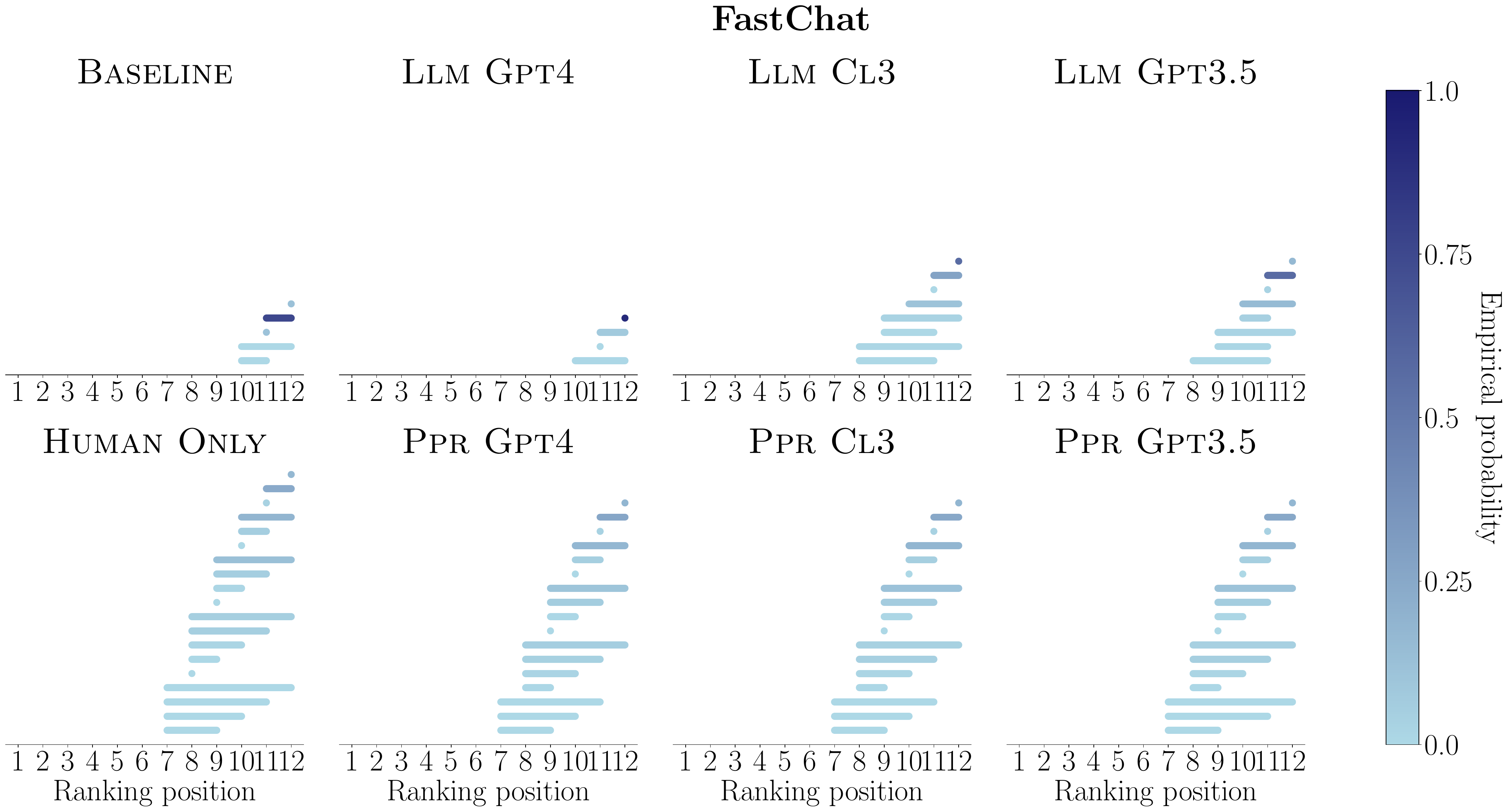}}\\
\captionsetup{list=off,format=cont}
\caption{}
\label{fig:all-ranksets}
\end{figure}

\newpage
\section{Synthetic Experiments}
\label{app:synthetic}

In this section, we evaluate our framework in a synthetic setting where the true rank-sets of the models are known. This allows us to validate the theoretical coverage guarantee of Theorem~\ref{theorem:coverage} without relying on a proxy metric, and also compute the rank-biased overlap (RBO)~\cite{webber2010similarity}.

\xhdr{Experimental setup}
We consider $k=8$ models and, in each experiment, we generate a random vector of true win probabilities $\boldsymbol{\theta}$ (Eq.~\ref{eq:theta-star}), which induces the true rank-sets of the models.\footnote{In our experiments, we generate true win probabilities $\boldsymbol{\theta}$ with unique values, so the rank-sets are always singletons, resulting in a unique true rank for each model.} To obtain win probabilities $\breve{\boldsymbol{\theta}}$ (Eq.~\ref{eq:theta-model}), we add random noise to the vector and then re-normalize $\breve{\boldsymbol{\theta}}$. We sample the noise from a $\text{Uniform}(-u,u)$ distribution, where $u$ a parameter we manually set  to simulate different alignment levels between strong LLMs and human preferences. A larger $u$ indicates a greater difference between $\boldsymbol{\theta}$ and $\breve{\boldsymbol{\theta}}$, meaning the LLM is less aligned with human preference, and vice versa. In our experiments, we set $u\in\{0.05,0.1,0.3\}$ to simulate three different strong LLMs.

To draw reliable conclusions, for each experiment, we create rank-sets $300$ times, each time using a different set of $n+N=50{,}000$ simulated pairwise comparisons by humans and the three strong LLMs, with an equal number of pairwise comparisons per pair of models. We ensure that each model provides the first and second response to an equal number of pairwise comparisons.
Let $m_a$ and $m_b$ be the two models in a pairwise comparison, with $m_a$ giving the first response. For each pairwise comparison, first, we generate a number $x \in (0,1)$ uniformly at random. For the human outcome, if $x<2\,\theta_{m_a}$, then the response of model $m_a$ is preferred ($w=1$). Similarly, for the strong LLM outcome, if $x<2\,\breve{\theta}_{m_a}$, then the response of model $m_a$ is preferred ($\hat{w}=1$). In every comparison we set $w'=\hat{w}'=0$.

\begin{figure}[t]
\centering
\includegraphics[width=0.6\linewidth]{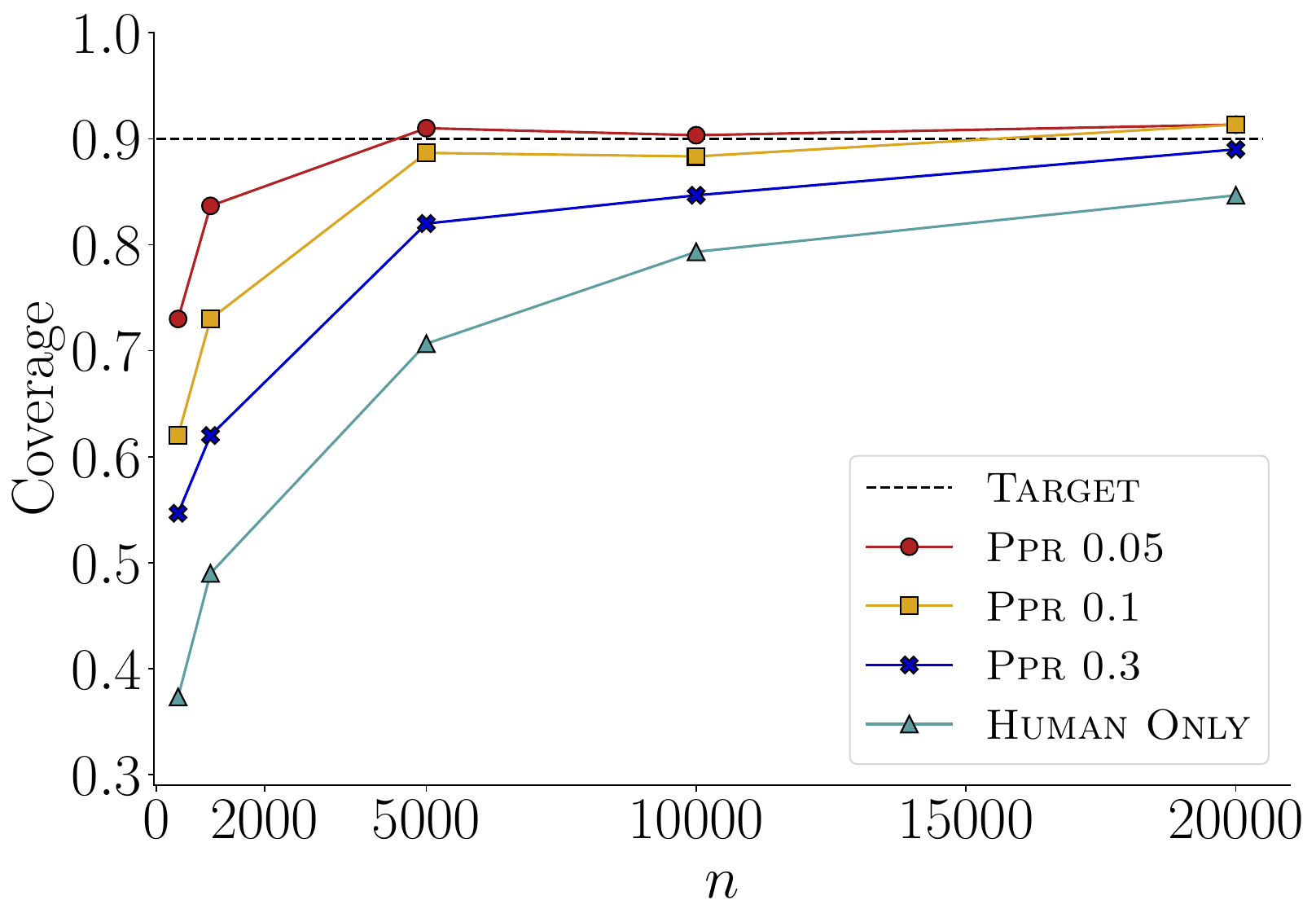}
\caption{
Empirical coverage probability of the rank-sets constructed using only $n$ synthetic pairwise comparisons by humans (\textsc{Human Only}) and using both $n$ synthetic pairwise comparisons by humans and $N+n$ synthetic pairwise comparisons by one out of three different simulated strong LLMs (\textsc{Ppr} $0.05$, \textsc{Ppr} $0.1$ and \textsc{Ppr} $0.3$) with $\alpha=0.1$ and $N+n=50000$.
Each of the strong LLMs has a different level of alignment with human preferences controlled by a noise value $u \in \{0.05, 0.1, 0.3\}$.
The dashed line indicates the $1-\alpha$ target coverage.
The empirical coverage of the rank-sets constructed using only $N+n$ synthetic pairwise comparison by one of the same three strong LLMs (not shown in the figure) is $0.38$ ($u=0.05$), $0.13$ ($u=0.1$) and $0.0$ ($u=0.3$).
} \label{fig:coverage-n-synthetic}
\end{figure}

\xhdr{Coverage probability}
Using the generated pairwise comparisons, we compute rank-sets in a similar way as described in Section~\ref{sec:experiments}, with $\alpha=0.1$. Since the true rank-sets are known, we can compute the (empirical) coverage probability, shown in Figure~\ref{fig:coverage-n-synthetic}. The results show that the coverage probability increases with $n$, consistent with Theorem~\ref{theorem:coverage}. Further, the coverage probability is greater when $u$ is smaller, indicating that our method achieves better results when the strong LLM is more aligned to human preference.

\begin{figure}[t]
\centering
\includegraphics[width=0.6\linewidth]{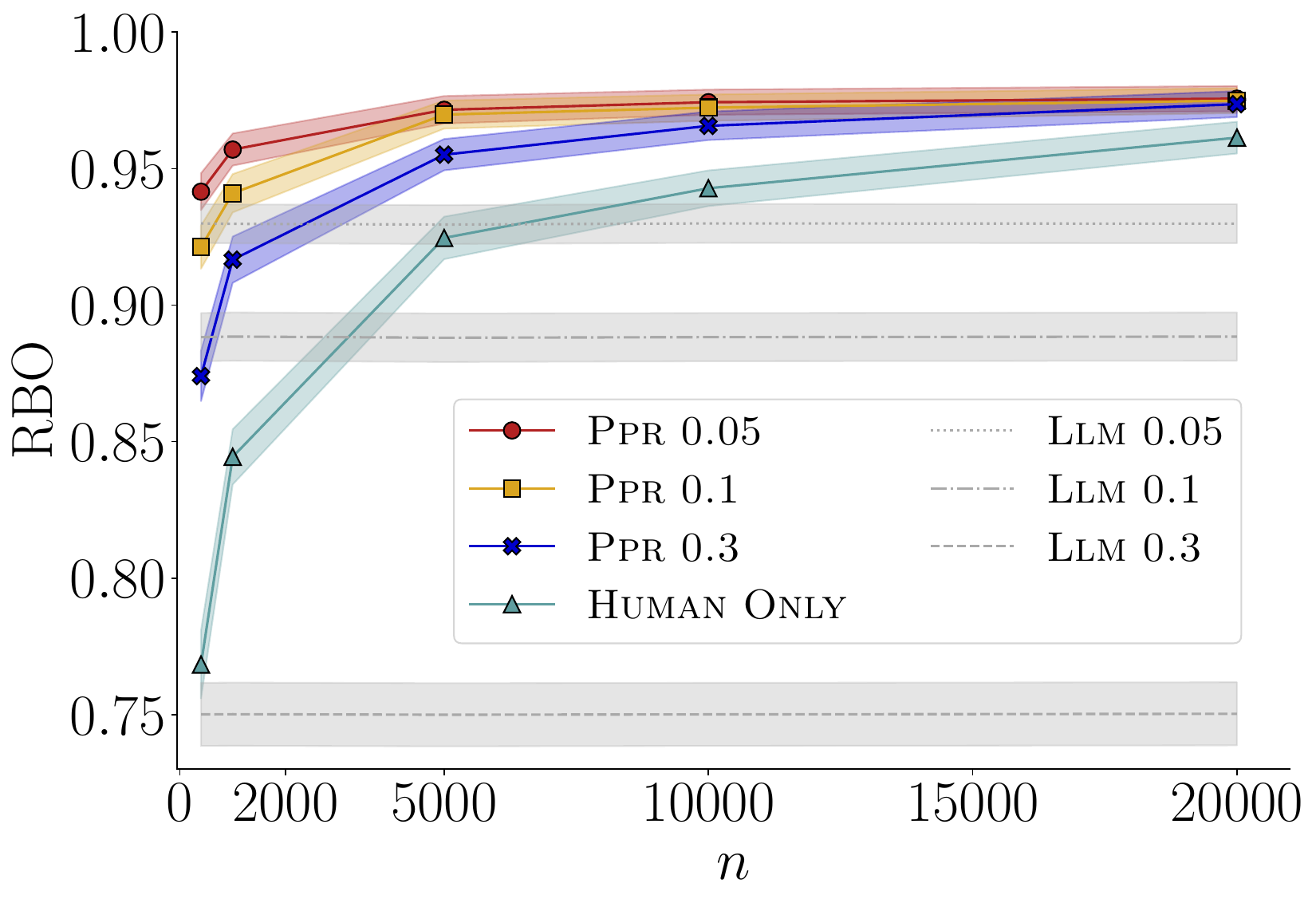}
\caption{
Average rank-biased overlap (RBO) of rankings constructed by ordering the empirical win probabilities $\boldsymbol{\hat{\theta}}$ estimated using only $N+n$ synthetic pairwise comparisons by one out of three different simulated strong LLMs (\textsc{Llm} $0.05$, \textsc{Llm} $0.1$ and \textsc{Llm} $0.3$), only $n$ synthetic pairwise comparisons by humans (\textsc{Human Only}), and both $n$ synthetic pairwise comparisons by humans and $N+n$ synthetic pairwise comparisons by one out of the same three strong LLMs (\textsc{Ppr} $0.05$, \textsc{Ppr} $0.1$ and \textsc{Ppr} $0.3$) for $\alpha=0.1$ and $N+n=50000$.
Each of the strong LLMs has a different level of alignment with human preferences controlled by a noise value $u \in \{0.05, 0.1, 0.3\}$.
RBO was computed with respect to the true ranking constructed by ordering the true win probabilities $\boldsymbol{\theta}$, for $p=0.6$.
The shaded region shows a 95\% confidence interval for the RBO among all 300 repetitions.
}
\label{fig:rbo-n}
\end{figure}

\xhdr{Rank-biased overlap (RBO)}
For each method, we obtain a ranking $\hat{T}$ by sorting the models in descending order of their $\hat{\boldsymbol{\theta}}$ values. We then compute the RBO metric relative to their true ranking $T$ as follows:
\begin{equation*}
    \text{RBO}(T,\That,p)=(1-p)\sum_{i\in[k]}{p^{i-1}\frac{|T_{:i}\cap \That_{:i}|}{i}}
\end{equation*}
where $T_{:i}$ and $\That_{:i}$ represents the top $i$ models in ranking $T$ and $\That$, respectively, and $p \in [0,1]$ is a chosen parameter. When $p=1$, all models are weighed equally. As we decrease $p$, more emphasis is given to the top-ranked models, and at $p = 0$, only the top-ranked model is considered.
In Figure~\ref{fig:rbo-n}, we compare RBO values as we increase the number of human pairwise comparisons $n$, for $p=0.6$. The results show that increasing $n$ improves RBO across all methods. Additionally, combining human pairwise preferences with a strong LLM further improves RBO values, demonstrating that our method performs better than those solely relying on strong LLM preferences. We repeated our experiments with multiple values of $p \in [0,1]$ and observed no significant variation in the results.

\end{document}